\documentclass[10pt,journal,compsoc]{IEEEtran}
\ifCLASSOPTIONcompsoc
  \usepackage[nocompress]{cite}
\else
  \usepackage{cite}

\fi

\usepackage[table]{xcolor}

\usepackage{hyperref}
\usepackage{url}
\usepackage{multirow}

\usepackage{graphicx}

\usepackage{algorithm}
\usepackage{algorithmic}
\usepackage{amsmath}
\usepackage{wrapfig}
\usepackage{amssymb}

\definecolor{minor}{RGB}{189, 30, 30} 

\usepackage{booktabs}
\usepackage{cleveref}

\begin{document}

\title{Improving Generalized Visual Grounding with Instance-aware Joint Learning}

\author{Ming~Dai,
        Wenxuan~Cheng,
        Jiang-Jiang~Liu,
        Lingfeng~Yang,
        Zhenhua~Feng,~\IEEEmembership{Senior~Member,~IEEE},
        Wankou~Yang,~\IEEEmembership{Member,~IEEE},
        and~Jingdong~Wang,~\IEEEmembership{Fellow,~IEEE}
\thanks{M. Dai, W. Cheng and W. Yang are with the School of Automation, Southeast University (emails: mingdai@seu.edu.cm; chengwenxuan@seu.edu.cn; wkyang@seu.edu.cn).}
\thanks{J.J Liu and J Wang are with Baidu Inc. (email: j04.liu@gmail.com; wangjingdong@outlook.com).}
\thanks{Z. Feng is with JiangNan University (email: fengzhenhua@jiangnan.edu.cn).}
\thanks{L. Yang is with Nanjing University of Science and Technology (email: yanglfnjust@njust.edu.cn).}
}

\IEEEtitleabstractindextext{
\begin{abstract}
    Generalized visual grounding tasks, including Generalized Referring Expression Comprehension (GREC) and Segmentation (GRES), extend the classical visual grounding paradigm by accommodating multi-target and non-target scenarios. Specifically, GREC focuses on accurately identifying all referential objects at the coarse bounding box level, while GRES aims for achieve fine-grained pixel-level perception.
    However, existing approaches typically treat these tasks independently, overlooking the benefits of jointly training GREC and GRES to ensure consistent multi-granularity predictions and streamline the overall process. Moreover, current methods often treat GRES as a semantic segmentation task, neglecting the crucial role of instance-aware capabilities and the necessity of ensuring consistent predictions between instance-level boxes and masks.
    To address these limitations, we propose \textbf{\textit{InstanceVG}}, a multi-task generalized visual grounding framework equipped with instance-aware capabilities, which leverages instance queries to unify the joint and consistency predictions of instance-level boxes and masks. 
    To the best of our knowledge, InstanceVG is the first framework to simultaneously tackle both GREC and GRES while incorporating instance-aware capabilities into generalized visual grounding.
    To instantiate the framework, we assign each instance query a prior reference point, which also serves as an additional basis for target matching. This design facilitates consistent predictions of points, boxes, and masks for the same instance.
    Extensive experiments obtained on ten datasets across four tasks demonstrate that InstanceVG achieves state-of-the-art performance, significantly surpassing the existing methods in various evaluation metrics. The code and model will be publicly available at \url{https://github.com/Dmmm1997/InstanceVG}.
\end{abstract}


}

\maketitle

\IEEEdisplaynontitleabstractindextext

%
\IEEEpeerreviewmaketitle

\IEEEraisesectionheading{\section{Introduction}\label{sec:introduction}}

%
%
%
%
\IEEEPARstart{C}{lassic} visual grounding aims to localize the referred target in an image based on a given textual sentence. It primarily includes Referring Expression Comprehension (REC)~\cite{mattnet,resc,dynamicmdetr} and Referring Expression Segmentation (RES)~\cite{cris,prompt-ris,simvg}. Specifically, REC focuses on perceiving the coarse-grained bounding box of the referred target, while RES requires identifying its fine-grained pixel-level mask.
These tasks utilize a free-form text description as the query, overcoming the constraints of restricted categories in conventional object detection~\cite{yolov3} and segmentation~\cite{unet} tasks, thereby enhancing both generality and usability.
A typical characteristic of the classic visual grounding tasks is the one-to-one relationship between the referring expression and target. Recently, generalized visual grounding has extended the classic paradigm by incorporating multi- and non-target scenarios. This extension enhances the rationality of visual grounding via a broader perspective, providing more reliable and adaptable algorithmic support for practical applications such as embodied AI~\cite{goat} and autonomous driving~\cite{planning}.

Existing REC methods can be broadly categorized into two-stage, one-stage, and transformer-based approaches. 
Two-stage methods~\cite{mattnet,rvg-tree,cm-att-erase,ref-nms} initially generates proposals using off-the-shelf detectors~\cite{fasterrcnn,maskrcnn} and then calculates the similarities between the referring expression and proposals, selecting the best match as the final prediction. 
One-stage methods~\cite{yang2019fast,resc} typically integrate language features with image feature maps and directly predict bounding boxes on dense grids with predefined anchors~\cite{yolov3}. 
Transformer-based approaches~\cite{transvg,mdetr,simvg} leverage the powerful contextual understanding capabilities of the self-attention mechanism~\cite{transformer} to facilitate image-text interactions.
Specifically, some existing methods~\cite{transvg,transvg++} employ direct regression techniques, while others~\cite{mdetr,dynamicmdetr} utilize the decoder architecture of DETR~\cite{detr} for prediction. 
Alternatively, the existing RES methods can be divided into cnn-based and transformer-based approaches. 
Previous methods~\cite{husegmentation2016, cmpc} typically rely on convolutional operations for cross-modal fusion to generate segmentation masks. 
Recent studies~\cite{lavt,vlt,mmm,cris,restr} leverage the global contextual interaction capabilities of self-attention mechanism~\cite{transformer} to enhance multimodal interaction.

\begin{figure}
  \centering
  \includegraphics[width=1.0\linewidth]{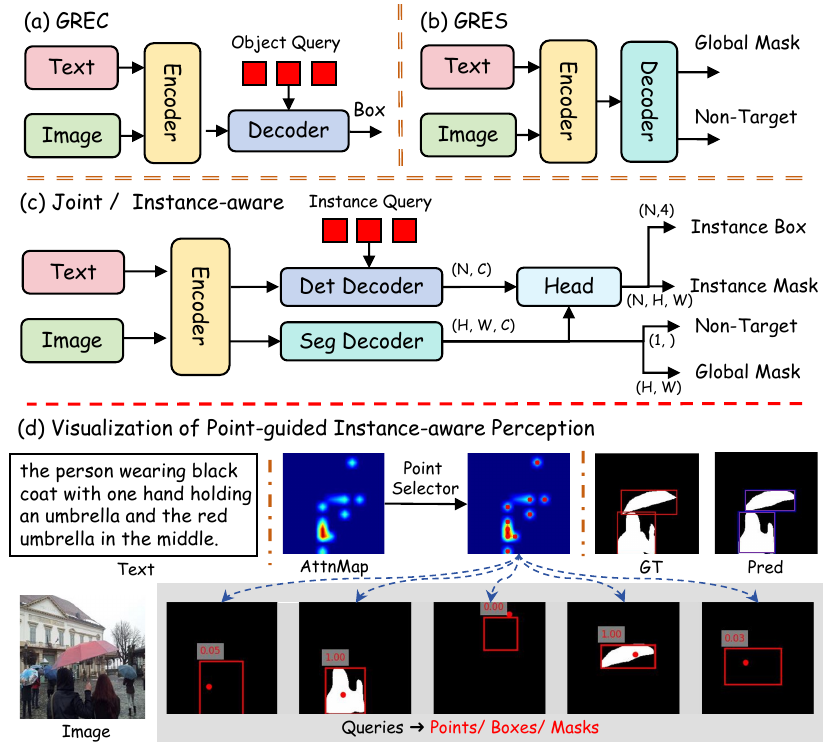}
  \vspace{-15pt}
  \caption{Comparison of generalized visual grounding tasks. (a) Transformer-based GREC paradigm, typically employing a DETR-style decoder for object localization; (b) Conventional GRES model architectures, adhering to a semantic segmentation paradigm and incorporating non-target predictions; (c) The proposed InstanceVG framework, which for the first time unifies GREC and GRES tasks within a query-based architecture; (d) InstanceVG leverages instance queries to seamlessly bridge detection and segmentation tasks. It adaptively filters prior points to provide instance queries with positional priors, ensuring consistent predictions across points, boxes, and masks. Here, we visualize the prior points selected by five representative instance queries, along with their corresponding predicted boxes and masks.
  }
  \label{fig:motivation}
\end{figure}

Although existing single-task models can achieve strong performance, deploying two separate models for detection and segmentation remains inefficient and inflexible for practical applications. To address this, MCN~\cite{mcn} introduced the first unified model for REC and RES, enforcing cross-task consistency via energy fields and post-processing strategies. 
Subsequent works~\cite{reftr,vg-law} explored multimodal fusion in multi-task settings, while others~\cite{seqtr,polyformer,pvd} reformulated visual grounding as a sequence prediction problem to generate the corner points of boxes and masks. 
However, these approaches are primarily designed for the \emph{classic} visual grounding setting, where each query corresponds to a single target instance, and thus fail to address the additional complexity of \emph{generalized} scenarios that involve multi-target and non-target cases. In such settings, prior methods~\cite{mcn,reftr,pvd,seqtr,C3VG} that directly regress a single bounding box become fundamentally inapplicable. 
Moreover, despite the growing interest in multi-task learning, the impact and feasibility of \emph{joint} training under generalized conditions remain largely unexplored. Existing frameworks also lack an explicit mechanism to ensure \emph{instance-aware} consistency between predicted boxes and masks, which is critical for robust grounding in complex scenes. 

In this work, we address these gaps by introducing a unified multi-task framework that explicitly incorporates instance queries to bridge detection and segmentation, enabling complementary and consistent predictions across tasks while simplifying the overall pipeline. Consequently, one of our primary research objectives is to investigate \textbf{\textit{(1) how to effectively perform multi-task joint training for generalized visual grounding, simplifying the pipeline while achieving complementary and instance-consistent predictions.}}

The exploration of multi-task architectures in generalized scenarios constitutes one of the core contributions of this work. The advent of generalized visual grounding, where a single referring expression may correspond to multiple target entities or none, naturally motivates the study of \emph{instance-aware} capabilities. In conventional single-modality object perception, instance segmentation is a well-established paradigm: methods such as Mask R-CNN~\cite{maskrcnn} and Mask2Former~\cite{mask2former} possess strong instance-level perceptual capabilities, yet they operate without constraints from referring language. In contrast, existing generalized referring expression segmentation (GRES) approaches~\cite{rela,hdc,DeRIS} often merge all instance masks into a single global mask for supervision, inherently disregarding the role of instance-level guidance in refining fine-grained perception. 

From a practical standpoint, when a referring expression targets multiple objects, the absence of instance-level segmentation makes it impossible to precisely localize and distinguish each entity, thereby diminishing the ability to provide accurate spatial relationship cues for downstream reasoning. Furthermore, instance-aware capabilities offer fine-grained supervision signals that strengthen the coupling between bounding boxes and masks, enabling the model to produce more robust and semantically expressive representations. 
To this end, we aim to address a second fundamental research question: \textbf{\textit{(2) how to design a principled framework that equips generalized visual grounding with instance-aware capabilities, leveraging fine-grained supervision to enhance referential understanding in complex, real-world scenarios.}}

To address the aforementioned issues, we propose an instance-aware multi-task architecture for generalized visual grounding, named \textbf{\textit{InstanceVG}}. Below, we discuss in detail how InstanceVG addresses the two key challenges.

\textbf{\textit{(1) Jointly training multi-task generalized visual grounding.}} 
Existing GREC~\cite{grec,simvg} methods, as illustrated in Fig.~\ref{fig:motivation}(a), adopt a DETR-type~\cite{detr,deformabledetr} query-based architecture, achieving target localization via one-to-one query-to-target matching. 
Current GRES methods~\cite{rela, groundhog, gsva, hdc}, as depicted in Fig.~\ref{fig:motivation}(b), typically employ two separate branches: one for predicting semantic masks and the other for determining the existence of referents. 
The proposed InstanceVG method, shown in Fig.~\ref{fig:motivation}(c), seamlessly integrates GREC and GRES through a query-guided architecture. 
A distinctive feature of InstanceVG is its ability to consistently predict both boxes and masks via instance queries. 
Unlike single-task detection, a significant challenge arises: \textit{how can we ensure that an instance query predicts both the bounding box and the instance mask for the same target?} In simple terms, this involves binding the query with the corresponding instance. To address this, we achieve consistent predictions by matching the predicted boxes with their corresponding masks during training, leveraging the Hungarian matching algorithm to establish correspondences among queries, boxes, and masks. 
Moreover, to further enhance the final segmentation performance, we jointly train the global semantic and instance-level segmentation branches, and fuse their outputs to generate the final prediction.

\textbf{\textit{(2) Query-based instance-aware perception.}}
Recent DETR-series~\cite{detr} studies have emphasized the design of object queries, including the integration of prior information~\cite{conditionaldetr,dabdetr} and the exploration of query matching strategies~\cite{dndetr, dino, groupdetr}. Additionally, several prior-based approaches~\cite{sam, ferret, dynamicmdetr} leveraging points as prompts have emerged, where most methods adopt dense grid settings or employ manually defined points interactively as prompts. 
In this paper, we propose a novel approach for adaptively selecting high-response points based on heatmap responses as prior reference points, with intermediate visualization shown in Fig.~\ref{fig:motivation}(d). 
The architecture of InstanceVG is depicted in Fig.~\ref{fig:framework}. 
First, we introduce an \textit{attention-based point-prior decoder} (APD), which adaptively identifies high-response points to serve as priors for instance queries. 
These points act as sampling references to interact with multi-scale image features via a deformable decoder, thereby enhancing the representation of instance queries. 
Subsequently, we propose a \textit{point-guided instance-aware perception head} (PIPH) to establish correspondence among points, boxes, and masks. 
Through interaction with semantic features, PIPH generates semantic queries for instance mask prediction. 
To enhance the overall referential semantic segmentation capability, we simultaneously train both the global semantic and instance-level segmentation branches, leveraging fine-grained capabilities to further improve referential perception performance.
As shown in Fig.~\ref{fig:motivation}(d), prior reference points are filtered based on attention distributions, directing the instance queries toward the nearest corresponding targets via point-guided object matching. The last row of Fig.~\ref{fig:motivation}(d) demonstrates the predicted bounding boxes and instance masks for five queries, along with their reference points.

To summarize, the main contributions of this paper are as follows:
\begin{itemize}
    \item We propose a query-guided multi-task architecture, named \textbf{InstanceVG}, which, for the first time, unifies the training of GREC and GRES tasks within a single framework. This not only simplifies the pipeline but also fosters complementary predictions.
    \item InstanceVG pioneers the introduction of instance-aware capabilities into generalized visual grounding, endowing the task with fine-grained instance-level predictions, thereby enhancing its flexibility in real-world applications. Additionally, by embedding fine-grained perception capabilities into the referential segmentation task, it further improves the robustness and adaptability of referential understanding.
    \item We design an attention-based point-prior decoder and a point-guided instance-aware perception head, which seamlessly connect the multi-task framework and establish consistency among points, bounding boxes, and instance masks. This design enhances the directivity and interpretability of instance queries.
    \item The proposed InstanceVG framework achieves state-of-the-art performance on the RefCOCO/+/g (REC/RES), gRefCOCO (GREC/GRES), Ref-ZOM, and R-RefCOCO/+/g datasets, delivering significant improvements over the existing methods.
\end{itemize}

The remainder of this paper is structured as follows: Sec.~\ref{sec:related_work} provides a comprehensive review of related work. 
Sec.~\ref{sec:method} details the proposed InstanceVG framework, including its architectural design and key components. 
In Sec.~\ref{sec:experiment}, we describe the experimental setup and present results that validate the effectiveness of our approach. 
Sec.~\ref{sec:visualization} showcases qualitative visualizations to further demonstrate the performance of InstanceVG. 
Last, Sec.~\ref{sec:conclusion} draws the conclusion and discusses potential directions for future research.

\section{Related Work}
\label{sec:related_work}
In this section, we review the existing literature relevant to our research. 
We categorize the related work into three main areas: referring expression comprehension/segmentation (Sec.~\ref{subsec:rec_res}), generalized referring expression comprehension/segmentation (Sec.~\ref{subsec:gres_grec}), and multi-task visual grounding (Sec.~\ref{subsec:mtvg}). 
We discuss the evolution of methods in each area, highlighting key advancements and identifying gaps that our proposed InstanceVG framework aims to address.

\subsection{Referring Expression Comprehension / Segmentation}
\label{subsec:rec_res}
In classical REC, a single sentence corresponds to one target bounding box. Early two-stage methods~\cite{mattnet,ddpn,rvg-tree,cm-att-erase,ref-nms} addressed this by first generating candidate proposals and subsequently matching the referring expression to these proposals. Later, one-stage methods~\cite{yang2019fast,realgin,mcn,resc} adopted a dense anchor strategy~\cite{yolov3}, enabling more efficient inference. In recent years, Transformer-based approaches~\cite{qrnet,seqtr,transvg,lads,transvg++,dynamicmdetr,simvg,oneref} have been developed to effectively model cross-modal relationships, offering significant improvements over earlier methods.  
On the other hand, RES is a task in which a single sentence corresponds to a set of pixels. 
Classical methods~\cite{husegmentation2016, liu2017, cmpc, efn} predominantly relied on convolution-based operations for cross-modal fusion to generate segmentation masks. 
To address the limitations in vision-language relationship modeling inherent in these approaches, recent works~\cite{lavt,vlt,mmm,cris,restr} have adopted advanced attention-based mechanisms~\cite{transformer} to enhance multimodal interactions.  
Among these, some approaches~\cite{simvg,oneref,PropVG} improve referential understanding by decoupling multimodal fusion from downstream tasks and repositioning it as an upstream pre-training process. 
Building upon this foundation, our approach leverages the powerful visual-text understanding capabilities of a multi-modality encoder~\cite{beit3} and extends it to the generalized instance-aware visual grounding setting. 
Furthermore, we design an adaptive point-guided perception architecture that seamlessly integrates multi-task learning with instance-aware reasoning, ensuring robust and coherent task performance.

\begin{figure*}
  \centering
  \includegraphics[width=1.0\linewidth]{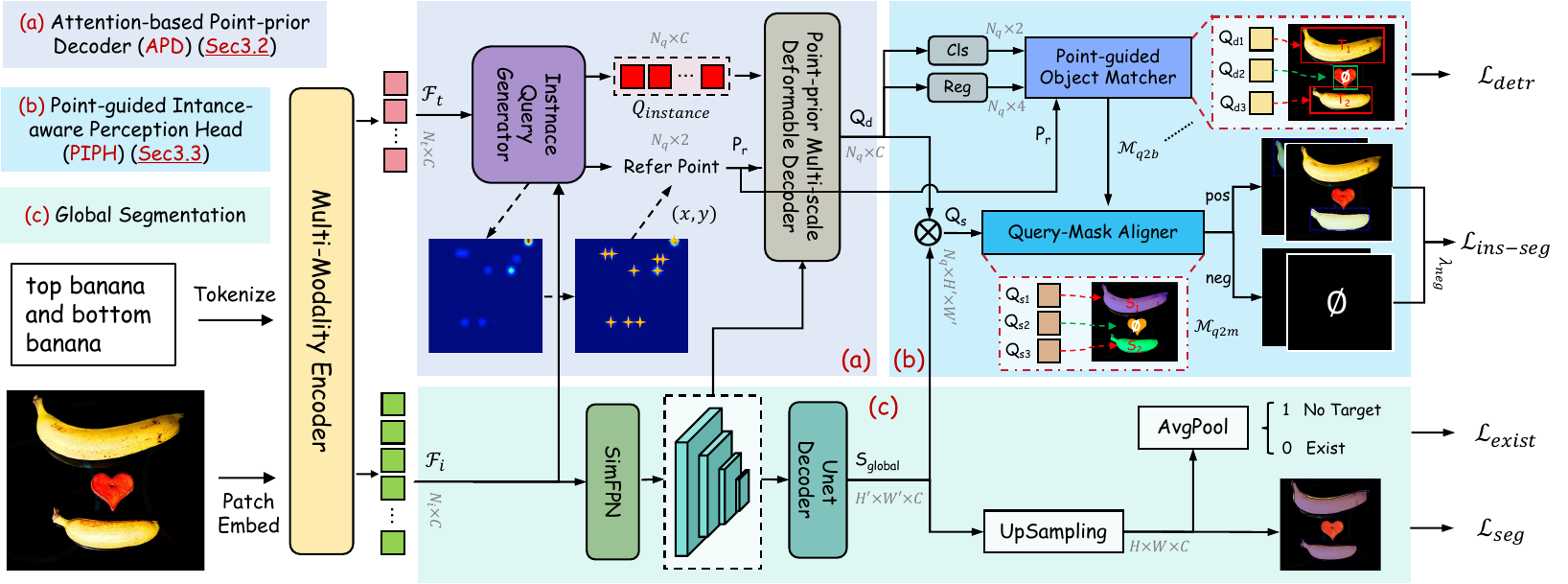}
  \vspace{-15pt}
  \caption{Overview of InstanceVG. First, the Multi-Modality Encoder simultaneously fuses the referring expression and image features. 
  The model is then divided into three key components: (a) and (b) form the instance prediction branch, which aims to achieve consistency and alignment between instance boxes and masks; (c) corresponds to the global segmentation module, which provides global semantic segmentation capabilities and determines the existence of target objects. 
  Specifically, (a) is the proposed Attention-based Point-prior Decoder (APD) module, which generates prior reference points and interacts with multi-scale semantic features to produce instance queries with target reasoning capabilities. 
  (b) is the proposed Point-guided Instance-aware Perception Head (PIPH), which establishes correspondences among queries, boxes, and masks, thereby enforcing consistency constraints across points, boxes, and masks.}
  \label{fig:framework}
\end{figure*}

\subsection{Generalized Referring Expression Comprehension / Segmentation}
\label{subsec:gres_grec}
Recently, to address the inflexibility of REC with one-to-one pairing, ReLA~\cite{rela} introduced the generalized RES task, which broadens the scope to include both non-target and multi-target scenarios. Furthermore, GREC~\cite{grec} extended GRES from segmentation to detection tasks. Similarly, DMMI~\cite{refzom} proposed a new benchmark for beyond-single-target segmentation, while RefSegformer~\cite{rris} enhanced transformer-based models with non-target discrimination, achieving robust segmentation performance. However, these methods predict a global semantic mask that aggregates all targets, neglecting the importance of fine-grained instance-level supervision.  
In contrast, this paper pioneeringly introduces fine-grained instance-level supervision into generalized visual grounding tasks. We propose a point-guided instance-aware perception head that establishes explicit correspondences between queries and objects or instances, enabling consistent predictions. Additionally, by integrating instance-level supervision with global semantic prediction, the proposed InstanceVG achieves enhanced perception robustness and fine-grained alignment in generalized scenarios.

\subsection{Multi-Task Visual Grounding} 
\label{subsec:mtvg}
Multi-task visual grounding aims to simultaneously localize and segment referring targets within a unified model. 
MCN~\cite{mcn} pioneered the joint training of REC and RES tasks, achieving task simplification and complementary benefits. Subsequently, Transformer-based approaches~\cite{reftr,vg-law,eevg, C3VG, PropVG} have explored advanced multimodal modeling techniques, further improving performance in multi-task visual grounding. Notably, several methods~\cite{seqtr,polyformer,pvd} employ sequential transformer architectures that integrate visual and textual data, iteratively refining predictions to enhance task effectiveness.
More recently, the research leveraging multimodal large language models (MLLMs)~\cite{llava, li2024survey} has extended the field by incorporating rule-based serialization strategies. 
These methods~\cite{kosmos, lisa, gsva, glamm, ferret} unify REC and RES tasks within a single framework, marking significant progress in multimodal understanding.
Despite these advancements, multi-task visual grounding under generalized scenarios remains underexplored. 
Thus, this paper introduces a novel framework that seamlessly integrates GREC and GRES tasks into a unified architecture.

\section{The Proposed InstanceVG Method}
\label{sec:method}
This section outlines the overall architecture of \textbf{InstanceVG} (Sec.~\ref{subsec:overview}). 
We introduce the attention-based point-prior decoder, which generates informative reference points integrated into object queries via the point-prior deformable decoder for multi-scale feature interaction (Sec.~\ref{subsec:apd}).  
Next, the point-guided instance-aware perception head is presented, ensuring consistent predictions across queries, objects, and instances (Sec.~\ref{subsec:piph}). We then describe the training strategy for robust task performance (Sec.~\ref{subsec:training_objectives}) and conclude with post-processing techniques (Sec.~\ref{subsec:postprocess}).

\subsection{Overview}
\label{subsec:overview}

\begin{figure}
  \centering
  \includegraphics[width=1.0\linewidth]{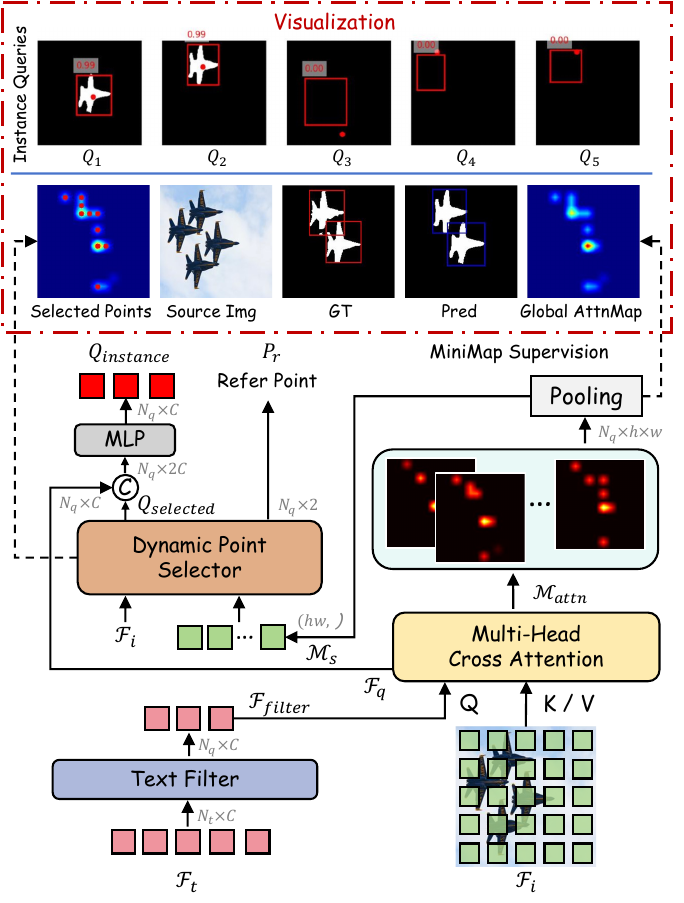}
  \vspace{-20pt}
  \caption{The IQG module is designed to select reference points and contextually enrich the instance queries. First, a text filter filters $K$ effective and highly responsive tokens from $\mathcal{F}_t$, which are then used as Q in the multi-head cross-attention mechanism, with $\mathcal{F}_i$ serving as the K and V. Next, the dynamic point selector employs a greedy algorithm to select reference points by balancing both spatial distance and response scores, ensuring these points comprehensively cover all referential instances. In the right panel, $Q_i$ represents the point, box, and mask prediction corresponding to the $i^\text{th}$ instance query.
  }
  \label{fig:aqsm}
\end{figure}

We first provide an overview of the InstanceVG architecture in Fig.~\ref{fig:framework}. The process begins with a multi-modality encoder~\cite{beit3}, which performs vision-language encoding and feature interaction by jointly processing an image $\mathcal{I} \in \mathbb{R}^{H \times W \times 3}$ and a textual expression $\mathcal{T}$. 
Specifically, the image $\mathcal{I}$ is transformed into patch embeddings, while the textual expression $\mathcal{T}$ is tokenized. 
The compressed features from both modalities are concatenated and fed into the BEiT-3~\cite{beit3} model for joint encoding.
For detailed processing steps, please refer to SimVG~\cite{simvg}. 
The encoder outputs are transformed back into image features and textual features.
Next, an image linear layer and a text linear layer are applied separately to project these features into a lower-dimensional space $C$, yielding $\mathcal{F}_i \in \mathbb{R}^{N_i \times C}$ and $\mathcal{F}_t \in \mathbb{R}^{N_t \times C}$. 
The architecture then branches into two core components. 
The first component builds upon the global segmentation branch (Fig.~\ref{fig:framework}(c)) commonly adopted by mainstream methods~\cite{rela,hdc}. 
Specifically, it employs SimFPN~\cite{vitdet} to extend the single-layer output of the ViT~\cite{vit} backbone into multi-scale feature representations. 
Subsequently, a simple U-Net~\cite{unet} decoder is utilized to integrate hierarchical information, producing global semantic segmentation predictions $S_{global} \in \mathbb{R}^{H' \times W' \times C}$, where $H'=\frac{H}{4}$ and $W'=\frac{W}{4}$. The global segmentation component serves four purposes: (1) providing global semantic segmentation predictions; (2) determining the existence of referents; (3) providing multi-scale image features for APD (Fig.~\ref{fig:framework}(a)); and (4) interacting with the decoded queries $Q_d$ in PIPH (Fig.~\ref{fig:framework}(b)) to generate instance-aware semantic queries $Q_s$.
The second component is the instance perception branch, which predicts both instance boxes and instance masks. 
It comprises the APD (Fig.~\ref{fig:framework}(a)) and the PIPH (Fig.~\ref{fig:framework}(b)). 
APD serves two primary functions: (1) adaptively filtering prior points that extensively cover potential referring targets based on attention responses; and (2) injecting these prior points into instance queries, enabling multi-scale interactions with image features via a deformable decoder. 
PIPH, on the other hand, establishes correspondences among prior reference points, instance boxes, and instance masks. For instance segmentation, it leverages the dot product between the decoded queries $Q_d$ and the global semantic feature to obtain instance-aware semantic queries, which ultimately guide instance-aware perception.

\subsection{The Attention-based Point-prior Decoder}
\label{subsec:apd}
The structure of APD is illustrated in Fig.~\ref{fig:framework}(a), comprising two primary components: the Instance Query Generator (IQG) module and the point-prior multi-scale deformable decoder. Initially, IQG adaptively generates the instance queries $Q_\text{instance} \in \mathbb{R}^{N_q \times C}$ and their associated prior points $P_r \in \mathbb{R}^{N_q \times 2}$.  
Subsequently, the deformable decoder dynamically retrieves contextual information from multi-scale image features to generate decoded queries $Q_d \in \mathbb{R}^{N_q \times C}$, encapsulating information from diverse targets.  
In the following sections, we detail the architecture of the IQG module in Sec.~\ref{subsubsec:aqg} and elaborate on the point-prior multi-scale deformable decoder in Sec.~\ref{subsubsec:ppmdd}.

\subsubsection{Instance Query Generator}
\label{subsubsec:aqg}

IQG aims to adaptively select suitable initial reference points and query embeddings. 
To this end, as shown in Fig.~\ref{fig:aqsm}, we first use a text filter to filter $N_q$ effective queries from the $N_t$ text tokens, which are then used as Q in cross-attention with image features (K/V). 
The attention map is obtained by:
\begin{equation}
    \mathcal{M}_{\text{attn}} = \text{Softmax} \left( \frac{\mathcal{F}_{\text{filter}} \cdot \mathcal{F}_i^{\top}}{\sqrt{d_k}} \right) \cdot \mathcal{F}_i.
\end{equation}

Next, we perform average pooling on $\mathcal{M}_{\text{attn}}$ across the $N_q$ channels to obtain the spatial score distribution map $\mathcal{M}_s=\text{Mean}(\mathcal{M}_{\text{attn}}, \text{dim}=0)$. 
Then, we employ a dynamic point selector to adaptively select prior location points $P_r$ from $\mathcal{F}_i$ that cover all possible referential instances and their corresponding queries $Q_\text{selected}$. By concatenating $Q_\text{selected}$ with $\mathcal{F}_q$ and passing them through an MLP layer, we obtain the instance query embeddings $Q_\text{instance}$. This process can be expressed as follows:
\begin{equation}
    Q_\text{instance} = \text{MLP}( \text{Concat}( Q_\text{selected}, \mathcal{F}_q)).
\end{equation}


\begin{algorithm}
    \caption{Text Selector}
    \label{alg:STS}
    \footnotesize
    \begin{algorithmic}[1]
    \REQUIRE Text feature set $\mathcal{F}_t\in \mathbb{R}^{N_t \times C}$, text valid mask $m_t \in \{0,1\}^{N_t}$, query selection number $N_q$
    \ENSURE Filtered feature set $\mathcal{F}_{\text{filter}} \in \mathbb{R}^{N_q \times C}$
    \STATE Mask valid features: $\mathcal{F}_{\text{valid}} = \mathcal{F}_t \odot m_t$
    \STATE Compute L2 norm scores: $s = \|\mathcal{F}_{\text{valid}}\|_2$
    \STATE Count valid features: $V = \sum m_t$
    \IF{$V \geq N_q$}
        \STATE Select top-$N_q$ features: $\mathcal{F}_{\text{filter}} = \text{TopK}(\mathcal{F}_{\text{valid}}, N_q)$
        \STATE Set mask: $m_{\text{filter}} = \mathbf{1}^{N_q}$
    \ELSE
        \STATE Select all valid features: $\mathcal{F}_{\text{filter}} = \mathcal{F}_{\text{valid}}$
        \STATE Pad to $N_q$: $\mathcal{F}_{\text{filter}} \leftarrow \text{Pad}(\mathcal{F}_{\text{filter}}, N_q)$
        \STATE Set mask: $m_{\text{filter}} = \text{Pad}(m_t, N_q)$
    \ENDIF
    \RETURN $\mathcal{F}_{\text{filter}}$
    \end{algorithmic}
\end{algorithm}
\normalsize 


\noindent{\bf Text Filter} selects effective and highly responsive tokens from the text token set. This process balances two main factors. 
First, it excludes padding tokens, retaining only the tokens containing meaningful information. Second, it evaluates each token's responsiveness using an L2 norm score. This strategy preferentially selects the tokens with high scores and valid content. The details for text filter are described in Algorithm~\ref{alg:STS}. 
The primary function of the text filter is to select a specified number of high-response features from a feature set based on a given mask. 
First, the algorithm filters the valid features using the mask and calculates their L2 norm scores. 
Then, it compares the number of valid features with the predefined selection number \( N_q \). If the number of valid features is greater than or equal to \( N_q \), the top \( N_q \) features with the highest scores are selected, and the corresponding mask is set to all ones. 
Otherwise, all valid features are selected, and padding is applied to reach \( N_q \) features, with the mask being filled accordingly. Last, the algorithm returns the selected feature set and the corresponding mask. 
The selection of \( N_q \) is analyzed in Table~\ref{table:ablation_num_query}. By default, we set \( N_q \) to 10.

\begin{algorithm}
\caption{Dynamic Point Selector}
\label{alg:GPS}
\footnotesize
\begin{algorithmic}[1]
\REQUIRE Input attnmap $\mathcal{M}_s \in \mathbb{R}^{h \times w}$, num of points $N_q$, distance weight $W_{\text{dist}}$, image features $\mathcal{F}_i$
\ENSURE Selected points $P_r \in \mathbb{R}^{N_q}$, Filtered queries ${Q}_\text{filter} \in \mathbb{R}^{N_q\times C}$
\STATE Apply sigmoid: $\mathcal{M}_s \leftarrow \sigma(\mathcal{M}_s)$
\STATE Initialize points set ${P}_r$
\STATE Candidate points: $P = \{(i, j) \mid i \in [1, h], j \in [1, w]\}$
\STATE Find max point: $p_{\text{max}} = \arg\max(\mathcal{M}_s)$
\STATE Add $p_{\text{max}}$ to $P_r$ and remove from $P$
\FOR{$k = 1$ to $N_q-1$}
    \STATE Compute minimum distance $\mathcal{D}$ from each point in $P$ to any point in $P_r$
    \STATE Compute combined score: $S = \mathcal{M}_s + W_{\text{dist}} \times \mathcal{D}$
    \STATE Select best point: $p_{\text{best}} = \arg\max({S})$
    \STATE Add $p_{\text{best}}$ to $P_r$ and remove from $P$
\ENDFOR
\STATE Select corresponding image features from $\mathcal{F}_i$ to generate ${Q}_{\text{filter}}$.
\RETURN $P_r$, ${Q}_\text{selected}$
\end{algorithmic}
\end{algorithm}
\normalsize 

\noindent{\bf Dynamic Point Selector} 
ensures that the selected points are not overly concentrated on a few specific instances. 
Instead, the selection aims to cover as many potential target instances as possible. 
Dynamic point selector employs a greedy algorithm, as detailed in Algorithm~\ref{alg:GPS}. 
The selection criterion prioritizes points with high scores while maintaining maximum distance from previously selected points, as outlined in line 8 of Algorithm~\ref{alg:GPS}. 
Last, based on the selected set of points $\text{R} = \left\{ (x_i, y_i) \mid i = 1, 2, \dots, N_q \right\},$ the corresponding queries $Q_\text{selected}$ are extracted from $\mathcal{F}_i$. 
By default, $W_\text{dist}$ is set to 0.003.

\begin{figure}
    \centering
    \includegraphics[width=0.75\linewidth]{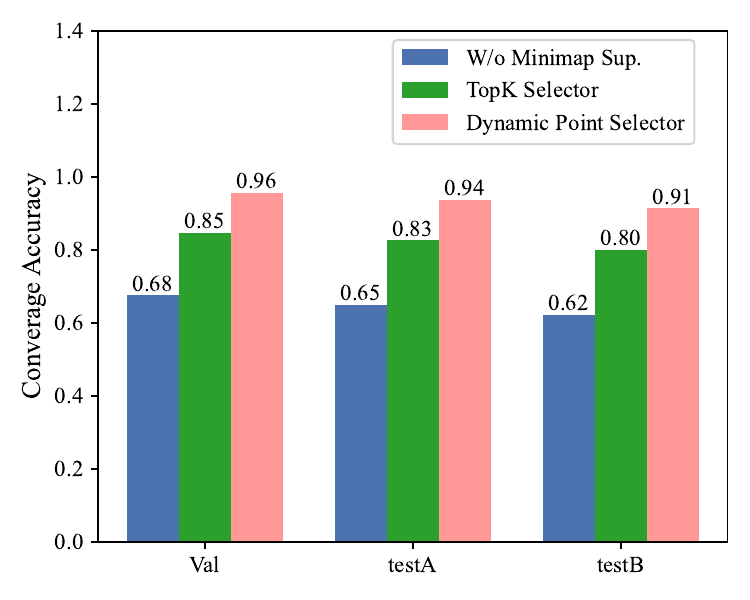}  
    \vspace{-10pt}
    \caption{\textbf{Bar chart of CoverAcc.} The `W/o Minimap Sup.' bar represents the results without supervision on the attention map. The `TopK Selector' bar indicates the use of the TopK strategy to select $N_q$ queries.
    }
    \label{fig:coveracc}
\end{figure}

To rigorously express the advantages of dynamic point selector, we introduce a metric called coverage accuracy, which measures the quality of points based on their coverage of targets:
\begin{equation}
    \text{CoverAcc} = \frac{1}{N}\sum\frac{\text{TP}}{\text{TP}+\text{FN}},
\end{equation}
where TP denotes the number of targets covered by points within the target box regions, and FN represents the number of targets not covered by any point. 
The denominator reflects the total number of targets. 
From Fig.~\ref{fig:coveracc}, we can draw two conclusions. 
First, supervision from minimap enhances the reliability of point selection in the attention map. 
Second, dynamic point selector significantly improves the coverage of instances by reference points.

\subsubsection{Point-prior Multi-scale Deformable Decoder}
\label{subsubsec:ppmdd}

Based on the ingenious design of deformable attention in Deformable DETR~\cite{deformabledetr}, we first review the computation of multi-scale deformable attention.
Given an input feature map $x \in \mathbb{R}^{C \times H_f \times W_f}$, let $q$ index a query element with content feature $z_q$ and a 2D reference point $p_q$. 
The multi-scale deformable attention is computed as follows:
\begin{equation}
  \begin{aligned}
  \text{MSDeAttn}(z_q, &p_q, \{x^l\}_{l=1}^{L})=\sum_{m=1}^{M} W_m \big[\sum_{l=1}^{L}\sum_{k=1}^{K} \\  
  & A_{mlqk}\cdot W'_m x^l(\phi_{l}(p_q) + \Delta p_{mlqk})\big],
  \label{eq:deform_attn_fun}
  \end{aligned}
  \end{equation}
  where $m$ indexes the attention head and $M$ represents the total number of attention heads, $k$ indexes the sampled keys, and $K$ represents the total number of sampled keys ($K \ll H_fW_f$). $l$ indexes the input feature level, $L$ refers to the total number of input feature levels.
  $\Delta p_{mlqk}$ and $A_{mlqk}$ denote the sampling offset and attention weight of the $k^\text{th}$ sampling point in the $m^\text{th}$ attention head, respectively. $z_q$ is the query feature, which will be used to calculate attention weights $A_{mlqk}$. Function $\phi_{l}(p_q)$ re-scales the coordinates $p_q$ to the input feature map of the $l^{\text{th}}$ level.
  $W'_m$ and $W_m$ are of learnable weights.

However, the initial sampling points in the multi-scale deformable decoder are typically densely defined on a grid, leading to the use of hundreds of object queries. 
Considering computational efficiency and the fact that the number of target objects in referring tasks rarely exceeds 10, we aim to reduce the number of object queries to decrease the decoder's computational overhead. 
To achieve this, we adopt the reference points generated by the IQG as prior locations for the initial sampling points in multi-scale deformable attention. 
Specifically, we replace $p_q$ with the generated reference points $p_r$, 
Additionally, the object query transitions from $z_q$ to the generated instance queries, $Q_\text{instance}$. 
Also, our point-prior multi-scale deformable decoder replaces the grid-based predefined $p_q$ with the adaptively selected points $p_r$, which provide each instance query with a point prior to the referent's approximate central location.
This design not only reduces computational complexity but also provides strong positional priors, thereby enhancing overall performance. 
The experimental analysis for this component is presented in Table~\ref{tab:ablation_decoding}.

\subsection{The Point-guided Instance-aware Perception Head}
\label{subsec:piph}

APD assigns each query a prior reference position for the target instance. To explicitly establish the correspondence among query, box, and mask, we design PIPH, as illustrated in Fig.~\ref{fig:framework}(b).  
PIPH addresses two key challenges: (1) how to interact with semantic features to construct instance semantic queries for instance-aware supervision; and (2) how to leverage the prior reference points of queries to identify the most relevant targets, thereby ensuring consistent predictions across points, boxes, and masks.
To tackle the first challenge, we compute a response mask $Q_s \in \mathbb{R}^{N_q \times H' \times W'}$ for each query by multiplying the decoded queries $Q_d$ with the semantic features $S_{global}$. This design enables the queries to extract richer global semantic information, effectively guiding instance-level segmentation.
For the second challenge, we propose a point-guided object matcher that introduces a cost function based on the distance between the reference point and the target's center, ensuring consistency between points and bounding boxes. Additionally, we define a query-mask aligner, which establishes correspondences between boxes and masks to further align points with instance masks. 
Last, we apply an instance-level segmentation loss to supervise both positive and negative masks, thereby enhancing instance-aware capability.

We now provide a detailed explanation of how to establish the correspondence between points, boxes, and masks, and how to implement consistency prediction. This process consists of two main operations: the point-guided object matcher (Sec.~\ref{subsubsec:point-guided-object-matcher}) and the query-mask aligner (Sec.~\ref{subsubsec:instance-aware-consistency-relation}).

\subsubsection{Point-guided Object Matcher}
\label{subsubsec:point-guided-object-matcher}
The point-guided object matcher utilizes reference points to guide the matching process between the decoded instance queries $Q_d$ and the targets. 
A key feature of this module is that, in contrast to the traditional bipartite graph matching used in DETR~\cite{detr}, it not only considers the predicted bounding boxes and categories of the queries but also introduces the prior point $P_r$ to influence the matching process. 
The goal is to bring $P_r$ as close as possible to the center of target, thereby enhancing the guiding role of the prior information.
In the specific implementation, we introduce an additional weighting term in the cost matrix that accounts for the distance between the reference point and the center of the target bounding box. 
This modification establishes a more direct association between the reference points and the targets. The cost is defined as:
\begin{equation}
\begin{aligned}
    C_{ij} & = \lambda_{\text{cls}} \text{CE}(p_i^{\text{cls}}, p_j^{\text{cls}}) 
             + \lambda_{\text{box}} \text{L}_1(p_i^{\text{box}}, p_j^{\text{box}}) \\
             & + \lambda_{\text{giou}} \text{GIoU}(p_i^{\text{box}}, p_j^{\text{box}}) 
             + \lambda_{\text{point}} \text{L}_1(p_i^{\text{point}}, p_j^{\text{center}}),
\end{aligned}
\label{eq:cost}
\end{equation}

where, \( p_i \) represents the \( i^\text{th} \) instance query, while \( p_j \) denotes the \( j^\text{th} \) ground truth target instance. Specifically, \( p_i^{\text{cls}} \) is the predicted foreground score, which is compared with the ground truth class \( p_j^{\text{cls}} \) using a cross-entropy loss to compute the classification cost. Similarly, \( p_i^{\text{box}} \), the predicted bounding box, is compared with the ground truth box \( p_j^{\text{box}} \) using both L1 loss and GIoU loss to derive the localization cost. 
In particular, \( p_i^{\text{point}} \), the reference point for the \( i^\text{th} \) instance query, is compared with the center of the ground truth box \( p_j^{\text{center}} \) using L1 loss to compute the point-based cost. 
By default, the loss weights are set as follows: \( \lambda_{\text{cls}} = 1.0 \), \( \lambda_{\text{box}} = 5.0 \), and \( \lambda_{\text{giou}} = 2.0 \), consistent with the original DETR settings. 
Additionally, this paper introduces a new hyperparameter, \( \lambda_{\text{point}} = 2.0 \). 
The ablation study of \( \lambda_{\text{point}} \) is reported in Table~\ref{table:ablation_point_cost}.

\subsubsection{Query-Mask Aligner}
\label{subsubsec:instance-aware-consistency-relation}
After applying the point-guided object matcher, we obtain the query-to-object matching relationship $\mathcal{M}_{q2b}$. 
The query-mask aligner then propagates $\mathcal{M}_{q2b}$ to establish the query-to-mask matching relationship $\mathcal{M}_{q2m}$, leveraging the one-to-one correspondence between boxes and masks. 
This correspondence is feasible due to two key foundations: (1) At the data level, a one-to-one relationship between gt boxes and gt masks is accessible. (2) At the prediction level, the sequences of $Q_d$ and $Q_s$ are aligned in order. These default settings ensure the consistency among points, boxes, and masks can be effectively constructed.

\subsection{Training Objectives}
\label{subsec:training_objectives}

The training objective includes four parts.
(1) \textit{Detection}: The detection part employs a loss function $\mathcal{L}_{detr}$ similar to DETR~\cite{detr}, incorporating the L1, Cross-Entropy, and GIoU loss functions to handle the detection task.
(2) \textit{Semantic segmentation}: This part uses the BCE and Dice loss~\cite{mask2former} functions, similar to those used in~\cite{cris}, to quantify the difference between the gt and predicted global masks: $\mathcal{M}_{gt}$ and $S_{global}$.
(3) \textit{Instance-level segmentation}: This part also adopts the same BCE and Dice losses~\cite{mask2former} as used in the global semantic segmentation component.
During the matching process, a one-to-one correspondence between the semantic queries and their instance target is established, enabling the computation of the loss for positive sample pairs. However, without proper suppression, the predictions for negative samples become uncontrollable. 
To address this and make the query assignments more precise, we impose constraints on the sample masks. 
Additionally, a weighting factor \( \lambda_{\text{neg}} \) is applied to balance the loss contributions from negative samples.

(4) \textit{Non-target discrimination}: This branch is responsible for binary classification and employs the BCE loss to distinguish the existence of referents.
The overall training objective is formulated as:
\begin{equation}
\begin{aligned}
\mathcal{L}_\text{total}  = & \ \lambda_\text{detr} \cdot \mathcal{L}_\text{detr} + \lambda_\text{seg} \cdot \mathcal{L}_\text{seg} \\ 
& +  \lambda_\text{instance} \cdot \mathcal{L}_\text{ins-seg} + \lambda_\text{exist} \cdot \mathcal{L}_\text{exist},
\end{aligned}
\end{equation}
where the hyperparameters are set by default to \( \lambda_{\text{detr}} = 0.1 \), \( \lambda_{\text{seg}} = 1.0 \), \( \lambda_{\text{instance}} = 1.0 \), and \( \lambda_{\text{exist}} = 0.2 \).  
The instance-level segmentation loss \( \mathcal{L}_\text{ins-seg} \) is defined as:
\begin{equation}
\label{eq:insseg}
\begin{aligned}
\mathcal{L}_\text{ins-seg} = \frac{1}{N_\text{pos}} \sum_i \mathcal{L}_{\text{seg}}^{\text{pos}} 
+ \frac{\lambda_\text{neg}}{N_\text{neg}} \sum_j \mathcal{L}_{\text{seg}}^{\text{neg}},
\end{aligned}
\end{equation}
where \( N_{\text{pos}} \) and \( N_{\text{neg}} \) denote the numbers of positive and negative sample masks, respectively.  
In this formulation, the contributions of positive and negative samples are weighted accordingly, with \( \lambda_{\text{neg}} \) set by default to \( 0.2 \).  

\begin{figure}
  \centering
  \includegraphics[width=1.0\linewidth]{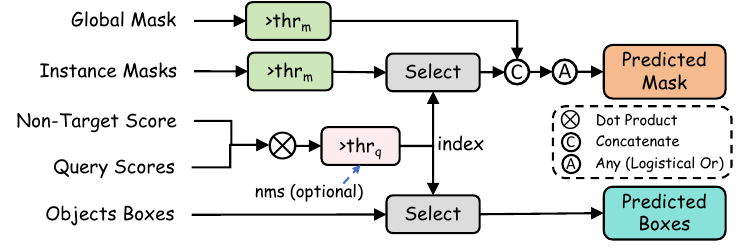}
  \caption{\textbf{Illustration of Post-processing}.
  First, both global and instance masks are binarized using $thr_m$. Then, we merge the non-target score and query scores through the dot product, and queries with scores greater than $thr_q$ are selected. Last, the instance and global mask are combined using a logistical OR operation.
  }
  \label{fig:postprocess}
\end{figure}

\subsection{Post-processing}
\label{subsec:postprocess}

Due to the introduction of instance masks, the post-processing in InstanceVG differs significantly from previous methods. The pipeline is illustrated in Fig.~\ref{fig:postprocess}. First, we apply weights to the query scores and the non-target score to reduce false positives in scenes without targets. A threshold $thr_q$ is used to identify valid queries, denoted as $index$. The detection branch then filters and outputs the corresponding targets based on these indices. For the segmentation branch, the global mask is combined with the instance masks. A threshold $thr_m$ is applied to select the pixel-level foreground mask. Last, the global mask is concatenated with the instance masks filtered by $index$, and a logical OR operation is applied to address incomplete instances.

\section{Experimental Results} 
\label{sec:experiment}

This section first introduces the used datasets in Sec.~\ref{subsec:dataset} and then describes the evaluation metrics in Sec.~\ref{subsec:metrics}. The experimental setups are outlined in Sec.~\ref{subsec:exp_setups}. Subsequently, we compare InstanceVG with existing state-of-the-art methods across 10 datasets spanning 4 tasks in Sec.~\ref{subsec:sota}. Last, Sec.~\ref{subsec:ablation} presents the ablation study results, analyzing the impact of different components and hyperparameters on the model's overall performance.

\begin{table*}
    \setlength{\tabcolsep}{6pt}
    \renewcommand{\arraystretch}{1.2}
    \centering
    \caption{
        Comparison with the state-of-the-art models on the RefCOCO/+/g~\cite{refcoco,refcocog-umd} datasets for REC task. `FT' indicates whether fine-tuning is performed on the target dataset. `Time' refers to the model inference latency per sample on a single GTX 1080Ti with a batch size of 1. Precision@0.5 is adopted as the evaluation metric.
    }
    \vspace{-10pt}
    \resizebox{0.87\linewidth}{!}
    {
        \begin{tabular}{l|c|c|c|ccc|ccc|cc|c}
        \specialrule{.1em}{.05em}{.05em}
        \multirow{2}{*}{Models} & \multirow{2}{*}{Visual Encoder} & \multirow{2}{*}{FT} & {Pre-train} & \multicolumn{3}{c|}{RefCOCO} & \multicolumn{3}{c|}{RefCOCO+} & \multicolumn{2}{c|}{RefCOCOg} & Time \\
        & & & Images & val & testA & testB & val & testA & testB & val-u & test-u & (ms)\\
        \hline
        \multicolumn{12}{c}{\textit{MLLM Methods}} \\
        \hline
        Shikra-7B~\cite{shikra} & CLIP-ViT-L & \checkmark& 0.5M & 87.01 & 90.61 & 80.24 & 81.60 & 87.36 & 72.12 & 82.27 & 82.19 & -\\
        Ferret-7B~\cite{ferret} & CLIP-ViT-L & \checkmark& $>$8M & 87.49 & 91.35 & 82.45 & 80.78 & 87.38 & 73.14 & 83.93 & 84.76 & -\\
        LION-4B~\cite{lion} & EVA-GPT &\checkmark &  3.6M & 89.73 & 92.29 & 84.82 & 83.60 & 88.72 & 77.34 & 85.69 & 85.63 & -\\
        \hline
        \multicolumn{12}{c}{\textit{Specialist Methods}} \\
        \hline
        MDETR~\cite{mdetr} & ResNet-101 &\checkmark & 200K & 86.75 & 89.58 & 81.41 & 79.52 & 84.09 & 70.62 & 81.64 & 80.89 & 108 \\
        SeqTR~\cite{seqtr} & DarkNet-53 & \checkmark& 174K & 87.00 & 90.15 & 83.59 & 78.69 &84.51 &71.87 & 82.69& 83.37  &  \textbf{50} \\
        UniTAB~\cite{unitab} &ResNet-101 & \checkmark& 200K&  88.59 & 91.06 & 83.75 & 80.97 & 85.36 &71.55& 84.58 &84.70 &  - \\
        DQ-DETR~\cite{dqdetr} & ResNet-101 & \checkmark& 6.5M & 88.63 & 91.04 & 83.51 & 81.66 &86.15 &73.21 & 82.76& 83.44 &  - \\
        GroundingDINO~\cite{groundingdino} & Swin-T & \checkmark& 7.2M & 89.19 &  91.86 & 85.99 & 81.09 &87.40 &74.71 & 84.15& 84.94& 120 \\
        PolyFormer~\cite{polyformer} & Swin-B &\checkmark & 174K & 89.73 &  91.73 & 86.03 & 83.73 &88.60 &76.38 & 84.46& 84.96&  152 \\
        PolyFormer~\cite{polyformer} & Swin-L & \checkmark& 174K & 90.38 &  92.89 & 87.16 & 84.98 &89.77 &77.97 & 85.83& 85.91 & -\\
        OFA-L~\cite{ofa} & ResNet-152 & \checkmark& 20M & 90.05 &  92.93 & 85.26 & 85.80 &89.87 &79.22 & 85.89& 86.55 & - \\
        \hline
        SimVG-B~\cite{simvg} & BEiT3-ViT-B & \checkmark& 174K & {91.47} & {93.65}  & {87.94 }& {84.83} & {88.85} &{79.12} & {86.30}& {87.26} & \underline{52} \\
        SimVG-L~\cite{simvg} & BEiT3-ViT-L & \checkmark& 28K & {92.93} & {94.70}  & \underline{90.28}& {87.28} & {91.64} &{82.41} & \underline{87.99}& \underline{89.15} & {116} \\
        OneRef-B~\cite{oneref} & BEiT3-ViT-B  & $\times$ & 0.5M & 89.16 & 92.03 & 87.26 & 83.18 &88.56 &77.66& 84.72& 85.17& - \\ 
        OneRef-B~\cite{oneref} & BEiT3-ViT-B & \checkmark& 0.5M & {91.89} & {94.31}  & {88.58}& {86.38} & {90.38} &{79.47} & {86.82}& {87.32} & - \\
        OneRef-L~\cite{oneref} & BEiT3-ViT-L & \checkmark& 0.5M & \underline{93.21} & \underline{95.43}  & 90.11& \underline{88.35} & \underline{92.11} &\underline{82.70} & {87.81}& {88.83} & - \\
        \hline
        \rowcolor{gray!10} \textbf{InstanceVG (Ours)} & BEiT3-ViT-B & $\times$ & 28K & {92.38} & {95.10} & {88.81} & {87.14} & {90.81} & {80.70} & {87.21} & {88.14} &  79 \\
        \rowcolor{gray!10} \textbf{InstanceVG (Ours)} & BEiT3-ViT-L & $\times$ & 28K & \textbf{94.42} & \textbf{96.04} & \textbf{92.39} & \textbf{90.12} & \textbf{92.89} & \textbf{85.94} & \textbf{89.58} & \textbf{90.62} & 130  \\
        \specialrule{.1em}{.05em}{.05em}
        \end{tabular}
    }
\label{table:sotarec}
\end{table*}

\begin{table*}
	\renewcommand{\arraystretch}{1.2}
	\setlength{\tabcolsep}{6pt}
    \centering
        \caption{Comparison with the state-of-the-art methods on RefCOCO/+/g~\cite{refcoco,refcocog-umd} datasets for RES task. `MT' indicates whether a multi-task paradigm is used for both REC and RES tasks. We abbreviate the datasets as follows: RefCOCO (RefC), ADE20K~\cite{ade20k} (A), COCO-Stuff~\cite{coco-stuff} (CS), PACO-IVIS~\cite{paco} (PL), PASCALPart~\cite{pascalpart} (PP), GranD~\cite{glamm} (G), and gRefCOCO~\cite{rela} (gRefC). mIoU is adopted as the evaluation metric.}
    \vspace{-10pt}
    \resizebox{0.98\linewidth}{!}{
	\begin{tabular}{l|c|c|c|ccc|ccc|cc|c}
		\specialrule{.1em}{.05em}{.05em}
		\multicolumn{1}{c|}{\multirow{2}{*}{Method}} &\multicolumn{1}{c|}{\multirow{2}{*}{Visual Encoder}} &  \multirow{2}{*}{MT} & Pre-train &
		\multicolumn{3}{c|}{RefCOCO} & \multicolumn{3}{c|}{RefCOCO+} & \multicolumn{2}{c|}{RefCOCOg} & \multicolumn{1}{c}{\multirow{1}{*}{Time}}\\
		\cline{5-12}
		& &  &  Data & val & test A & test B & val & test A & test B & val(U) & test(U) & (ms) \\
        \hline
        \multicolumn{13}{c}{\textit{MLLM Methods}} \\
        \hline
        LISA-7B~\cite{lisa} & SAM-ViT-H & $\times$ & A,CS,RefC,PL,PP&74.90 & 79.10 & 72.30 & 65.10 & 70.80 & 58.10 & 67.90 & 70.60 & - \\
		GSVA-7B~\cite{gsva} & SAM-ViT-H & $\times$ & A,CS,RefC,PL,PP,gRefC &77.20 & 78.90 & 73.50 & 65.90& 69.60 &59.80 &72.70 & 73.30 & - \\
        GLaMM-7B~\cite{glamm} & CLIP-ViT-H & $\times$ & G, RefC & 79.50 & 83.20& 76.90 & 72.60 & 78.70& 64.60& 74.20& 74.90 & - \\
        \hline
        \multicolumn{13}{c}{\textit{Specialist Methods}} \\
        \hline
		CRIS~\cite{cris} & ResNet101 &  $\times$& - &  70.47 & 73.18 & 66.10 & 62.27 & 68.06 & 53.68 & 59.87 & 60.36 & - \\
		LAVT~\cite{lavt} & Swin-B & $\times$& - & {74.46} & {76.89} & {70.94} & {65.81} & {70.97} & {59.23} & {63.34} & {63.62}& 135 \\
		ReLA~\cite{rela} & Swin-B & $\times$& - & 73.82 & 76.48 & 70.18 & 66.04 & 71.02 & 57.65 & 65.00 & 65.97 & - \\
		Prompt-RIS~\cite{prompt-ris} & CLIP-ViT-B & $\times$ &  RefC & 78.10 & {81.21} & 74.64 & 71.13& 76.60 & 64.25 & 70.47 & 71.29 & - \\
        OneRef-B~\cite{oneref} & BEiT3-ViT-B & $\times$& RefC & {79.83} & {81.86}  & {76.99}& {74.68} & {77.90} &{69.58} & {74.06}& {74.92} & - \\
        OneRef-L~\cite{oneref} & BEiT3-ViT-L & $\times$& RefC & {81.26} & {83.06}  & {79.45}& {76.60} & {80.16} & {72.95} & {75.68}& {76.82} & - \\
        \hline
		MCN~\cite{mcn} & DarkNet53 & \checkmark & - & 62.44 & 64.20 & 59.71 & 50.62 & 54.99 & 44.69 & 49.22 & 49.40 & \underline{56} \\
		SeqTR~\cite{seqtr} & DarkNet53 & \checkmark& RefC & 71.70 & 73.31 & 69.82 & 63.04 & 66.73 & 58.97 & 64.69 & 65.74 & \textbf{50}\\
		PolyFormer-B~\cite{polyformer} & Swin-B &\checkmark & RefC & 75.96 & 77.09 & 73.22 & 70.65 & 74.51 & 64.64 & 69.36 &69.88 & 152\\
        PolyFormer-L~\cite{polyformer} & Swin-L &\checkmark & RefC & 76.94 & 78.49 & 74.83 & 72.15 & 75.71 & 66.73 & 71.15 &71.17 & - \\
		PVD~\cite{pvd} & Swin-B & \checkmark& RefC & 74.82 & 77.11 &  69.52 & 63.38 & 68.60 & 56.92 & 63.13 & 63.62& - \\
        EEVG~\cite{eevg} & ViT-B & \checkmark& RefC & {79.49} & 80.87 & {77.39} & {71.86} & {76.67} & {66.31} &{73.56} & {73.47} & 117 \\
        PropVG~\cite{PropVG} & BEiT3-ViT-B & $\times$ & RefC & {81.96} & 83.58 & {80.02} & {77.14} & {79.83} & {72.18} &{76.97} & {77.72} & 76 \\
        DeRIS-L~\cite{DeRIS} & BEiT3-ViT-L & $\times$ & RefC & \underline{85.72} & \underline{86.64} & \underline{84.52} & \underline{81.28} & \underline{83.74} & \underline{78.59} &\underline{80.01} & \underline{81.32} & - \\
		\hline
		\rowcolor{gray!10}  \textbf{InstanceVG (Ours)} & BEiT3-ViT-B  & \checkmark & RefC &{81.36} & {83.05} & {79.28} & {76.64} & {79.51} & {71.56} & {75.89} & {76.59} & 79 \\
        \rowcolor{gray!10}  \textbf{InstanceVG (Ours)} & BEiT3-ViT-L  &\checkmark & RefC & \textbf{86.27} & \textbf{87.12} & \textbf{85.30} & \textbf{82.50} & \textbf{84.33} & \textbf{79.15} & \textbf{81.39} & \textbf{82.27} & 130 \\
		\specialrule{.1em}{.05em}{.05em}
	\end{tabular}}
	\label{tab:sota_res}
\end{table*}

\begin{table*}
\setlength{\tabcolsep}{6pt}
\centering
\renewcommand\arraystretch{1.2}
    \caption{Comparison with the state-of-the-art methods on gRefCOCO~\cite{rela} dataset for GRES task.}
    \vspace{-10pt}
    \resizebox{0.8\linewidth}{!}{
    \begin{tabular}{l|c|ccc|ccc|ccc}
    \specialrule{.1em}{.05em}{.05em}
    \multirow{2}{*}{Method} & \multirow{2}{*}{Backbone} & \multicolumn{3}{c|}{Val} & \multicolumn{3}{c|}{TestA}  & \multicolumn{3}{c}{TestB} \\
    & & gIoU & cIoU & N-acc. & gIoU & cIoU & N-acc. & gIoU & cIoU & N-acc.\\
    \hline
    \multicolumn{11}{c}{\textit{MLLM Methods}} \\
    \hline
    LISA-7B~\cite{lisa}& SAM-ViT-H & 61.63 & 61.76 & 54.67 & 66.27 & 68.50 & 50.01 & 58.84 & 60.63 & 51.91 \\
    GSVA-7B~\cite{gsva}& SAM-ViT-H & 66.47 & 63.29 & 62.43 & 71.08 & 69.93 & 65.31 & 62.23 & 60.47 & 60.56 \\
    
    \hline
    \multicolumn{11}{c}{\textit{Specialist Methods}} \\
    \hline
    MattNet~\cite{mattnet} & ResNet-101 & 48.24 & 47.51 & 41.15 & 59.30 & 58.66 & 44.04 & 46.14 & 45.33 & 41.32 \\
    LTS~\cite{lts} & DarkNet-53 & 52.70 & 52.30 & - & 62.64 & 61.87 & - & 50.42 & 49.96 & - \\
    VLT~\cite{vlt} & DarkNet-53 & 52.00 & 52.51 & 47.17 & 63.20 & 62.19 & 48.74 & 50.88 & 50.52 & 47.82 \\
    CRIS~\cite{cris} & CLIP-R101 & 56.27 & 55.34 & - & 63.42 & 63.82 & - & 51.79 & 51.04  & - \\
    LAVT~\cite{lavt} & Swin-B & 58.40 & 57.64 & 49.32 & 65.90 & 65.32 & 49.25 & 55.83 & 55.04 & 48.46 \\
    ReLA~\cite{rela} & Swin-B & 63.60 & 62.42 & 56.37 & 70.03 & 69.26 & 59.02 & 61.02 & 59.88 & 58.40 \\
    COHD~\cite{hdc} & Swin-B & {68.42} & {65.17} & {63.68} & {72.67} & {71.85} & {64.00} & {63.60} & {62.63} & {60.37} \\

    PropVG~\cite{PropVG} & BEiT3-ViT-B & 73.29 & \textbf{69.23} & 72.83 & \underline{74.43} & \underline{74.20} & 69.87 & \underline{65.87} & \underline{64.76} & \underline{64.97} \\
    DeRIS~\cite{DeRIS} & Swin-S + BEiT3-ViT-B & \textbf{74.10} & 68.06 & \textbf{77.03} & 73.72 & 71.99 & \textbf{75.98} & 65.63 & 64.65 & 63.44 \\
    \hline
    \rowcolor{gray!10} 
    \textbf{InstanceVG (Ours)} & BEiT3-ViT-B & \underline{73.36} & \underline{69.22} & \underline{72.84} & \textbf{75.21} & \textbf{74.51} & \underline{71.09} & \textbf{66.74} & \textbf{65.67} & \textbf{65.18} \\
    \specialrule{.1em}{.05em}{.05em}
    \end{tabular}
}
\vspace{-5pt}
\label{tab:sota_gres}
\end{table*}

\begin{table}
\centering
\setlength{\tabcolsep}{6pt}
\renewcommand{\arraystretch}{1.2}
\centering
\caption{Comparison with state-of-the-art methods on the Ref-ZOM~\cite{refzom} dataset.}
\vspace{-5pt}
\resizebox{0.9\linewidth}{!}{
    \begin{tabular}{l|c|ccc}
        \specialrule{.1em}{.05em}{.05em}
        {Method}& {Backbone} & oIoU & mIoU & Acc. \\ 
        \hline
        \multicolumn{5}{c}{\textit{MLLM Methods}} \\
        \hline
        LISA-7B~\cite{lisa}& SAM-ViT-H & 65.39 & 66.41 & 93.39 \\
        GSVA-7B~\cite{gsva}& SAM-ViT-H & 68.13 & 68.29 & 94.59\\
        \hline
        \multicolumn{5}{c}{\textit{Specialist Methods}} \\
        \hline
        MCN~\cite{mcn} & DarkNet-53 & 54.70 & 55.03 & 75.81 \\ 
        VLT~\cite{vlt} & DarkNet-53 & 60.43 & 60.21 & 79.26 \\ 
        LAVT~\cite{lavt} & Swin-B & 64.78 & 64.45 & 83.11 \\ 
        DMMI~\cite{refzom} & Swin-B & 68.21 & 68.77 & 87.02 \\
        CoHD~\cite{hdc} & Swin-B & \underline{69.81} & \underline{68.99} & \underline{93.34}\\
        \hline
        \rowcolor{gray!10} \textbf{InstanceVG (Ours)} & BEiT3-ViT-B & \textbf{71.52} & \textbf{71.12} & \textbf{97.42}\\
        \specialrule{.1em}{.05em}{.05em}
    \end{tabular}
}
\vspace{-10pt}
\label{table:sota_ref_zom}
\end{table}

\begin{table*}
\setlength{\tabcolsep}{6pt}
\renewcommand{\arraystretch}{1.2}
\centering
\caption{Comparison with state-of-the-art methods on the R-RefCOCO/+/g~\cite{rris} dataset.}
\vspace{-10pt}
\resizebox{0.7\linewidth}{!}{
    \begin{tabular}{l|ccc|ccc|ccc}
        \specialrule{.1em}{.05em}{.05em}
        \multirow{2}{*}{Method} & \multicolumn{3}{c|}{R-RefCOCO}  & \multicolumn{3}{c|}{R-RefCOCO+} & \multicolumn{3}{c}{R-RefCOCOg} \\
        \cline{2-10}
        & mIoU & mRR & rIoU & mIoU & mRR & rIoU & mIoU & mRR & rIoU  \\
        \hline
        CRIS~\cite{cris}  & 43.58 & 76.62 & 29.01 & 32.13 & 72.67 & 21.42 & 27.82	& 74.47 &	14.60 \\
        EFN~\cite{efn}    & 58.33  & 64.64 & 32.53 & 37.74  & 77.12 & 24.24 & 32.53 & 75.33 & 19.44 \\
        VLT~\cite{vlt}    &  61.66  & 63.36 & 34.05 & 50.15 & 75.37 & 34.19 & 49.67 & 67.31 & 31.64\\
        LAVT~\cite{lavt}   & 69.59 & 58.25 & 36.20 & 56.99 & 73.45 & 36.98 & 59.52 &  61.60 & 34.91 \\
        LAVT+~\cite{lavt}     & 54.70 & 82.39 & 40.11 & 45.99 & 86.35 & 39.71 & 47.22 & 81.45 & 35.46 \\
        RefSegformer~\cite{rris} & 68.78 & 73.73 & 46.08 & 55.82 & 81.23 & 42.14 & 54.99 &  71.31 & 37.65 \\ 
        CoHD~\cite{hdc}& {74.16} & {84.27} & {53.61} & {64.59} & {87.49} & {49.07} & {63.56} & {82.68} & {42.16} \\
        PropVG~\cite{PropVG}& \underline{75.86} & \textbf{92.39} & \underline{62.34} & \underline{69.39} & \underline{94.48} & \underline{59.04} & \underline{69.20} & \textbf{92.88} & \textbf{55.09} \\
        \hline
        \rowcolor{gray!10} 
        \textbf{InstanceVG (Ours)} & \textbf{76.73} & \underline{92.15} & \textbf{62.41} & \textbf{69.73} & \textbf{94.63} & \textbf{59.13} & \textbf{70.16} & \underline{92.30} & \underline{54.36} \\
        \specialrule{.1em}{.05em}{.05em} 
    \end{tabular}
}
    \label{tab:sota_rrefcoco}
    \vspace{-10pt}
\end{table*}

\begin{table}
\setlength{\tabcolsep}{2pt}
\renewcommand{\arraystretch}{1.2}
\centering
\caption{Comparison with the state-of-the-art methods on gRefCOCO~\cite{grec} dataset for GREC tasks. The threshold is set to 0.7 for all the methods.}
\vspace{-5pt}
\resizebox{1.0\linewidth}{!}{
    \begin{tabular}{l|cc|cc|cc}
        \specialrule{.1em}{.05em}{.05em}
        \multirow{2}{*}{Methods}  & \multicolumn{2}{c|}{Val} & \multicolumn{2}{c|}{TestA} & \multicolumn{2}{c}{TestB} \\
        &F1score   & N-acc.  &F1score   &  N-acc.   &F1score    & N-acc.   \\
        \hline
        MCN~\cite{mcn}   & 28.0 &30.6  & 32.3 & 32.0 & 26.8 & 30.3 \\
        VLT~\cite{vlt}  & 36.6 & 35.2  & 40.2 & 34.1 & 30.2 & 32.5 \\
        MDETR~\cite{mdetr} & 42.7 & 36.3 & 50.0 & 34.5 & 36.5 & 31.0   \\
        UNINEXT~\cite{uninext}  &58.2  &  50.6 & 46.4 & 49.3 & 42.9 & 48.2 \\
        SimVG~\cite{simvg} & {62.1} & {54.7} & {64.6} & {57.2} & {54.8}& {57.2} \\
        PropVG~\cite{PropVG} & \underline{72.2} & \underline{72.8} & \underline{68.8} &\underline{69.9} & \underline{59.0}& \underline{65.0} \\
        \hline
        \rowcolor{gray!10} \textbf{InstanceVG (Ours)} &  \textbf{73.5} & \textbf{72.8} & \textbf{70.2} &\textbf{71.1} & \textbf{60.8} & \textbf{65.2}\\
        \specialrule{.1em}{.05em}{.05em} 
    \end{tabular}
}
\vspace{-10pt}
\label{table:sota_grec}
\end{table}


\subsection{Datasets}
\label{subsec:dataset}
For the traditional tasks (REC/RES), we use the combined RefCOCO~\cite{refcoco}, RefCOCO+~\cite{refcoco}, and RefCOCOg~\cite{refcocog-umd} datasets for training, and evaluate on each individual subset.
For the GREC task, we use the gRefCOCO~\cite{grec} dataset for both training and testing.
For the GRES task, we independently train and evaluate on the gRefCOCO~\cite{rela}, Ref-ZOM~\cite{refzom}, and R-RefCOCO/+/g datasets~\cite{rris}.

\noindent{\bf RefCOCO/+/g.} The RefCOCO~\cite{refcoco}, RefCOCO+~\cite{refcoco}, and RefCOCOg~\cite{refcocog} datasets are widely utilized benchmarks for REC. 
RefCOCO and RefCOCO+ comprise 142,209 and 141,564 expressions, respectively, corresponding to 50,000 objects across nearly 20,000 images. 
In RefCOCO, the test A subset features images with multiple people, while testB focuses on images containing multiple object instances. RefCOCO+ is considered more challenging than RefCOCO due to the exclusion of location-based terms in its referring expressions. RefCOCOg contains 85,474 expressions referring to 54,822 objects in 26,711 images. 
It is characterized by longer and more complex expressions (averaging 8.4 words).

\noindent{\bf gRefCOCO.} The gRefCOCO dataset~\cite{grec,rela} extends REC tasks by incorporating expressions that reference multiple or non-targets, amounting to a total of 278,232 expressions. Of these, 80,022 are multi-target expressions, while 32,202 correspond to non-target expressions. 
The dataset includes 60,287 distinct object instances spanning 19,994 images, which are divided into training, validation, testA, and testB subsets, adhering to the UNC partition scheme of RefCOCO~\cite{refcoco}.

\noindent{\bf Ref-ZOM.} The Ref-ZOM dataset~\cite{refzom}, derived from the COCO dataset~\cite{coco}, comprises 55,078 images annotated with 74,942 objects. The dataset is split into 43,749 images with 58,356 objects for training and 11,329 images with 16,586 objects for testing. 
The annotations are categorized into three distinct scenarios: \textit{one-to-zero}, \textit{one-to-one}, and \textit{one-to-many}, corresponding to the non-, single-, and multi-target cases in GRES.

\noindent{\bf R-RefCOCO/+/g.} The R-RefCOCO~\cite{rris} dataset includes three variants: R-RefCOCO, R-RefCOCO+, and R-RefCOCOg, all derived from the classical RES benchmark. 
The dataset introduces negative sentences into the training set in a 1:1 ratio with positive sentences to enhance robustness. 
For evaluation, only the validation set adheres to the UNC partition principle, as officially specified. 

\textit{It is worth noting that to acquire instance masks and multi-task data for training, we utilize the corresponding annotations from the COCO~\cite{coco} dataset. The reprocessed dataset will be open-sourced to support and accelerate future research endeavors.}

\subsection{Evaluation Metrics} 
\label{subsec:metrics}
\noindent{\bf Tranditional Tasks (REC/RES).}
For REC, we evaluate the performance using Precision@0.5. The prediction is deemed correct if its IOU with the ground-truth box is larger than 0.5. 
For RES, we use mIoU as the evaluation metric.

\noindent{\bf Generalized Tasks (GREC/GRES).}
For GREC, we adopt {Pr@}(F$_1$=1, IoU$\geq$0.5) and N-acc. as the primary evaluation metrics. Here, {Pr@}(F$_1$=1, IoU$\geq$0.5), abbreviated as F1score, evaluates the performance using an F1score of 1 with an IoU threshold of 0.5.
For GRES, we evaluate our model using the following metrics: gIoU, cIoU, and N-acc for gRefCOCO~\cite{rela}; Acc., oIoU and mIoU for Ref-ZOM~\cite{refzom}; and mIoU, mRR, and rIoU for R-RefCOCO~\cite{rris}. The gIoU measures the average IoU across all instances in an image, treating empty targets as true positives with an IoU of 1. The cIoU metric evaluates intersection versus union pixels. In Ref-ZOM, mIoU and oIoU represent the average IoU for referred objects and cIoU, respectively. 
For R-RefCOCO, rIoU measures segmentation quality by including negative sentences in the mIoU calculation. N-acc. in gRefCOCO and Acc. in Ref-ZOM represent the ratio of correctly classified empty-target expressions. mRR in R-RefCOCO computes the recognition rate for empty-target expressions. 

\subsection{Experimental Setup}
\label{subsec:exp_setups}
For generalized tasks (GREC/GRES), the maximum sentence length is limited to 50 words. The model is trained for a total of 10 epochs, with the learning rate reduced by a factor of 10 at the 7th epoch. For traditional tasks (REC/RES), the maximum sentence length is restricted to 20 words. The model is trained for 20 epochs in total, with the learning rate decayed similarly at the 15th epoch. 
For the SOTA experiments, the input images are resized to a default resolution of $320 \times 320$. 
For the ablation studies, the input resolution is reduced to $224 \times 224$. 
The initial learning rate is set to $5 \times 10^{-5}$ for the multi-modality encoder and $5 \times 10^{-4}$ for the remaining parameters. The Adam optimizer is utilized for full-precision optimization, and no additional EMA strategy is applied.
All experiments are conducted using dual NVIDIA RTX 4090 GPUs.

\subsection{Comparison with the State-of-the-Art Methods}
We compare the performance of InstanceVG with existing state-of-the-art methods across 10 datasets (RefCOCO/+/g~\cite{refcoco,refcocog-umd}, gRefCOCO/+/g~\cite{grec,rela}, Ref-ZOM~\cite{refzom}, RRefCOCO/+/g~\cite{rris}) for 4 tasks (REC, RES, GREC, GRES).
\label{subsec:sota}
\subsubsection{Main Results on REC} 
Our approach seamlessly extends to traditional REC tasks. In fact, REC is merely a special case of GREC, representing a one-to-one matching relationship.
Table~\ref{table:sotarec} presents the results of the proposed InstanceVG method on the widely used RefCOCO/+/g datasets. 
Compared to the recent state-of-the-art specialist method OneRef~\cite{oneref}, InstanceVG achieves higher accuracy while utilizing fewer training samples. 
Specifically, under the same BEiT3-ViT-L configuration, our method improves the average accuracy by 1.4\% on RefCOCO, 1.9\% on RefCOCO+, and 2.4\% on RefCOCOg.
Furthermore, when compared to SimVG~\cite{simvg}, which uses the same 28K training samples, InstanceVG achieves an average improvement of 1.6\% on RefCOCO, 2.5\% on RefCOCO+, and 1.5\% on RefCOCOg. 
Notably, these improvements are achieved without requiring additional fine-tuning.

\subsubsection{Main Results on RES} 

Similarly, InstanceVG can be seamlessly extended to the RES task, with experimental results presented in Table~\ref{tab:sota_res}. 
As a multi-task framework, InstanceVG demonstrates superior performance compared to the SOTA multi-task model EEVG~\cite{eevg}. Under the same ViT-B backbone setting, our method achieves an average improvement of 2.0\% mIoU on the RefCOCO dataset, 4.3\% on RefCOCO+, and 2.7\% on RefCOCOg. Moreover, our ViT-L variant outperforms the similarly scaled SOTA model OneRef~\cite{oneref}, achieving average mIoU improvements of 5.0\% on RefCOCO, 5.4\% on RefCOCO+, and 5.6\% on RefCOCOg. 
Furthermore, our approach retains significant advantages even when compared to larger models (e.g., LISA~\cite{lisa}, GSVA~\cite{gsva}, GLaMM~\cite{glamm}) pre-trained on more extensive datasets.

\subsubsection{Main Results on GRES} 
To evaluate the effectiveness of our approach in a generalized setting, we first conduct a comparative analysis with the existing specialized methods on the gRefCOCO~\cite{rela} dataset, as presented in Table~\ref{tab:sota_gres}. 
The results demonstrate that our method establishes new SOTA performance across all the metrics in three evaluation sets of the large-scale GRES benchmark. 
Notably, compared with the existing SOTA method CoHD~\cite{hdc}, InstanceVG surpasses it with significant improvements of $+4.9\%$, $+2.5\%$, and $+3.1\%$ in gIoU on the val, testA, and testB sets, respectively. 
Furthermore, we report our results on the Ref-ZOM benchmark~\cite{refzom} in Table~\ref{table:sota_ref_zom}. 
Our method consistently outperforms CoHD~\cite{hdc}, achieving $+4.1\%$ improvement in Accuracy, $+1.7\%$ in oIoU, and $+2.1\%$ in mIoU. It is worth highlighting that our approach even surpasses GSVA~\cite{gsva}, which leverages the capabilities of MLLM~\cite{llava}. 
In addition, we extend our evaluation to the R-RefCOCO/+/g datasets~\cite{rris}. As illustrated in Table~\ref{tab:sota_rrefcoco}, our method achieves substantial improvements of $+8.8\%$, $+10.0\%$, and $+12.2\%$ in rIoU for the R-RefCOCO/+/g datasets when compared to CoHD~\cite{hdc}.

\subsubsection{Main Results on GREC} 
In addition to performing general segmentation, our InstanceVG model is also capable of handling detection tasks. We evaluate the detection performance of InstanceVG on the GREC~\cite{grec} dataset and compare it with existing SOTA methods.
The results are presented in Table~\ref{table:sota_grec}. Furthermore, under the same score threshold of $0.7$, InstanceVG significantly outperforms the existing SOTA method SimVG~\cite{simvg} with the improvements of $+11.4\%$, $+5.6\%$, and $+6.0\%$ in F1score on the validation, testA, and testB sets, respectively.

\subsection{Ablation Study}
\label{subsec:ablation}
\subsubsection{Effectiveness of The Core Modules}
\label{subsubsec:ablation_modules}
The core contributions discussed in this paper include: (1) the impact of multi-task joint learning on generalized visual grounding; (2) the influence of the proposed APD; and (3) the effectiveness of the proposed PIPH. As shown in Table~\ref{tab:ablation_modules}, multi-task joint supervision positively contributes to task complementarity in generalized scenarios, leading to performance improvements in both the GREC and GRES benchmarks. 
Specifically, the F1score increases by 1.2\%, and the gIoU improves by 2.3\%. 
After incorporating the APD module to augment the queries with prior position information, the F1score improves by 2.0\%, and the gIoU increases by 1.9\%. 
Last, introducing PIPH to construct a joint training architecture for instance-level and semantic segmentation, along with the consistent prediction between points, boxes, and masks, results in further improvements of 3.0\% in F1score and 2.3\% in gIoU. 
The experiments presented here provide analysis from two perspectives. 
On one hand, they validate that joint training in generalized scenarios can indeed provide complementary benefits. On the other hand, the designs of the APD and PIPH modules complement the instance-aware capability. 
By guiding the prediction of boxes and masks through point-guided instance queries within the multi-task architecture, the overall performance are significantly improved.

\begin{table}
  \makeatletter\def\@captype{table}
  \centering
  \footnotesize
  \setlength{\tabcolsep}{2.0pt}
  \renewcommand\arraystretch{1.2}
  \caption{Effectiveness of the core modules. APD and PIPH correspond to (a) and (b) in Figure~\ref{fig:framework}, respectively.}
  \vspace{-10pt}
  \resizebox{0.8\linewidth}{!}{
  \begin{tabular}{ccc|cccc}
  \specialrule{.1em}{.05em}{.05em} 
  Multi-Task         & APD        & PIPH       & F1score & N-acc. & gIoU & cIoU \\
  \hline
             &            &           &65.98  & 66.13 & 65.03 & 64.77 \\
  \checkmark &            &           &67.13  & 69.98 & 67.33 & 64.90 \\
  \checkmark & \checkmark &           &69.17  & 70.31 & 69.24 & 66.01 \\
  \checkmark & \checkmark & \checkmark  & $\textbf{71.43}$ & $\textbf{75.87}$ & $\textbf{72.41}$ & $\textbf{67.39}$ \\
  \specialrule{.1em}{.05em}{.05em}
  \end{tabular}
  }
  \label{tab:ablation_modules}
  \end{table}
\begin{table}
  \makeatletter\def\@captype{table}
  \centering
  \footnotesize
  \renewcommand\arraystretch{1.2}
  \setlength{\tabcolsep}{2.0pt}
  \caption{Effectiveness of different components of IQG. It includes a text filter (TF), a dynamic point selector (DPS), and a text-injected query (TIQ).}
  \vspace{-10pt}
  \resizebox{0.7\linewidth}{!}{
  \begin{tabular}{ccc|cccc}
  \specialrule{.1em}{.05em}{.05em} 
  TF & DPS & TIQ & F1score & N-acc. & gIoU & cIoU \\
  \hline
  &  &  & 69.20 & 69.95 & 70.43 & 66.36\\
  \checkmark &  & & 69.09 & 70.80 & 70.51 & 66.16\\
  \checkmark & \checkmark & & 71.16 & 72.87 & 71.73 & $\textbf{67.40}$ \\
  \checkmark & \checkmark & \checkmark & $\textbf{71.43}$ & $\textbf{75.87}$ & $\textbf{72.41}$ & 67.39 \\
  \specialrule{.1em}{.05em}{.05em}
  \end{tabular}}
  \label{tab:ablation_AQGM}
  \end{table}

\input{tables/ablation}

\begin{table*}[t]
\begin{minipage}{0.29\textwidth}
\centering
\setlength{\tabcolsep}{5pt}
\renewcommand{\arraystretch}{1.3}
\caption{Impact of different ratios of point cost $\lambda_{point}$.}
\vspace{-10pt}
\begin{tabular}{c|cccc}
\specialrule{.1em}{.05em}{.05em}
     $\lambda_{point}$ & F1score & gIoU & cIoU  \\
     \hline
     1.0 & 70.37 & 71.53 & 66.95 \\
     2.0 & $\textbf{71.43}$ & $\textbf{72.41}$ & $\textbf{67.39}$ \\
     5.0 & 69.90 & 71.47 & 66.85 \\
     10.0& 69.28 & 71.31 & 66.82 \\
\specialrule{.1em}{.05em}{.05em}
\end{tabular}
\label{table:ablation_point_cost}
\end{minipage}%
\hspace{0.02\textwidth}
\begin{minipage}{0.35\textwidth}
\centering
\setlength{\tabcolsep}{5pt}
\renewcommand{\arraystretch}{1.2}
\vspace{-5pt}
\caption{Impact of the number of queries. `TS' refers to the training schedule, where 1$\times$ denotes training for 10 epochs and 2$\times$ represents training for 20 epochs.
}
\vspace{-10pt}
\begin{tabular}{c|c|cccc}
\specialrule{.1em}{.05em}{.05em}
     $N_q$ & TS & F1score & N-acc. & gIoU & cIoU  \\
     \hline
     3 & 1$\times$ &70.26 & 72.74 & 71.32 & 66.74 \\
     5 & 1$\times$ &$\textbf{71.60}$ & 73.85 & 71.55 & 66.87 \\
     10 & 1$\times$ &71.43 & $\textbf{75.87}$ & $\textbf{72.41}$ & 67.39 \\
     30 & 1$\times$ &68.55 & 71.90 & 71.22 & 66.63 \\
     30 & 2$\times$ &71.53 & 75.83 & 72.36 & $\textbf{67.43}$ \\
\specialrule{.1em}{.05em}{.05em}
\end{tabular}
\label{table:ablation_num_query}
\end{minipage}
\centering
\hspace{0.01\textwidth}
\begin{minipage}{0.28\textwidth}
     \centering
     \setlength{\tabcolsep}{5pt}
     \renewcommand{\arraystretch}{1.2}
     \caption{Impact of query filter threshold $thr_q$ in post-processing.}
     \vspace{-10pt}
     \begin{tabular}{c|ccc}
     \specialrule{.1em}{.05em}{.05em}
           & F1score & gIoU & cIoU  \\
          \hline
          0.70 & 71.43 & 72.41 & 67.39 \\
          0.80 & 73.22 & 73.82 & $\textbf{68.17}$ \\
          0.85 & 74.12 & 74.37 & 68.15 \\
          0.90 & $\textbf{74.38}$ & $\textbf{74.59}$ & 67.58 \\
          0.95 & 73.53 & 73.79 & 65.62 \\
     \specialrule{.1em}{.05em}{.05em}
     \end{tabular}
     \label{table:ablation_threshold}
     \end{minipage}%
\end{table*}

\begin{table*}
\centering
\begin{minipage}{0.25\textwidth}
\centering
\setlength{\tabcolsep}{4pt}
\renewcommand{\arraystretch}{1.2}
\caption{Impact of the output type of mask in post-processing.}
\vspace{-10pt}
\begin{tabular}{l|cc}
\specialrule{.1em}{.05em}{.05em}
     Mask Output    & gIoU  & cIoU  \\
     \hline
     Only Global    & 72.61 & 65.66 \\
     Only Instance  & 74.19 & 67.18 \\
     Merge          & $\textbf{74.65}$ & $\textbf{67.66}$ \\
\specialrule{.1em}{.05em}{.05em}
\end{tabular}
\label{table:ablation_mask_output}
\end{minipage}
\hspace{0.01\textwidth}
\begin{minipage}{0.34\textwidth}
\centering
\setlength{\tabcolsep}{4pt}
\renewcommand{\arraystretch}{1.2}
\caption{Impact of non-target weighting and NMS in post-processing.}
\vspace{-10pt}
\begin{tabular}{cc|ccc}
\specialrule{.1em}{.05em}{.05em}
     NT Weighted & NMS  & F1score & gIoU & cIoU  \\
     \hline
                    &             & 73.18 & 73.94 & 67.20 \\
     \checkmark &             & 74.38 & $\textbf{74.59}$ & $\textbf{67.58}$ \\
     \checkmark & \checkmark  & $\textbf{74.71}$ & 74.55 & 67.48 \\
\specialrule{.1em}{.05em}{.05em}
\end{tabular}
\label{table:ablation_nms}
\end{minipage}
\hspace{0.01\textwidth}
\begin{minipage}{0.35\textwidth}
     \centering
     \setlength{\tabcolsep}{4pt}
     \renewcommand{\arraystretch}{1.2}
     \vspace{-10pt}
     \caption{Parameters of different core components.}
     \vspace{-10pt}
     \begin{tabular}{l|cc}
     \specialrule{.1em}{.05em}{.05em}
          Method  & Params (M) & MACs(G)  \\
          \hline
          BEiT-3   & 170.89 & 36.10 \\
          +SimFPN & 174.31(+3.42)& 40.02(+3.92)  \\
          +APD & 177.25(+2.94) & 43.69(+3.67) \\
          +UNet Decoder & 182.56(+5.31) & 58.55(+14.86) \\
          +PIPH & 182.69(+0.13) & 59.42(+1.13)\\
     \specialrule{.1em}{.05em}{.05em}
     \end{tabular}
     \label{table:params}
     \end{minipage}

\end{table*}

\subsubsection{Effectiveness of APD}
\label{subsubsec:ablation_apd}

\noindent{\bf Analysis of IQG.} 
The text filter chooses $N_q$ highly responsive tokens from the $N_t$ tokens of $\mathcal{F}_t$, thereby reducing the computational cost of multi-head cross attention. 
As observed in Table~\ref{tab:ablation_AQGM}, the introduction of the text filter results in almost no accuracy loss. 
Then, the dynamic point selector selects $N_q$ prior reference points covering different instances based on attention responses. Compared with the baseline which uses a strategy of selecting the top $N_q$ points, our dynamic point selector improves F1score by 2.1\% and gIoU by 1.2\%. 
Last, ITQ is designed to assist the learning of attention maps. 
It not only helps to optimize the attention map but also injects text information into the initial query, resulting in a +3.0\% improvement in N-acc and a +0.7\% increase in gIoU.

\noindent{\bf Analysis of Decoders.} 
As shown in Table~\ref{tab:ablation_decoding}, our baseline method employs the original DETR~\cite{detr} decoder. 
By incorporating the Deformable DETR~\cite{deformabledetr} decoder, we observe an improvement of 1.3\% in F1score and 0.5\% in gIoU. Furthermore, we utilize SimFPN to generate multi-scale features, and by adopting a hierarchical multi-scale Deformable DETR decoder, N-acc increases by 1.9\%. 
Last, leveraging the points filtered by the instance query generator module as the reference points for the Deformable DETR decoder yields an additional improvement of 0.5\% in F1score and 0.9\% in gIoU.

\subsubsection{Effectiveness of PIPH}

\noindent{\bf Analysis of Different Levels of Supervision.} 
As shown in Table~\ref{tab:ablation_IAH}, the impact of different levels of semantic supervision on performance is significant. Instance-level supervision alone outperforms global semantic supervision, enhancing both detection and segmentation performance by equipping the model with instance awareness. 
Interestingly, we find that the joint training with both global semantic and instance-level segmentation improves performance even without fusion during post-processing. 
We hypothesize that this is due to global supervision encouraging $S_{\text{global}}$ to produce strong semantic representations, thereby enhancing the discriminability of the semantic query mask $Q_s$. 
Last, we introduce the negative sample supervision, which guides the model to suppress mask predictions for negative sample queries. This additional supervision enhances the controllability of negative sample predictions, further improving segmentation performance.

\noindent{\bf Analysis of The Points Cost Weight}
In the point-guided object matcher, we introduce an additional point cost to the original DETR cost function.
The proportion of the point cost in the overall cost function determines the influence of the point prior on the query-target correspondence. 
To evaluate its impact, we conducted ablation studies on the weight of the point cost, $\lambda_{point}$, as presented in Table~\ref{table:ablation_point_cost}. 
The experimental results reveal that a small $\lambda_{point}$ diminishes the effect of the point prior, while a large $\lambda_{point}$ severely disrupts the box and classification predictions. 
Ultimately, we select $\lambda_{point} = 2$ as the optimal value for achieving balanced performance.

\noindent{\bf Analysis of The Number of Prior Points.}
Since InstanceVG establishes a one-to-one correspondence between queries and reference points, and in the setting of APD in this paper, the number of queries $N_q$ matches exactly with the number of prior points, it is crucial to study the impact of the $N_q$ on the results. We conducted experiments to explore the effect of the number of reference points on performance. 
As shown in Table~\ref{table:ablation_num_query}, increasing $N_q$ generally requires longer training times to achieve convergence. As the number of queries increases, the points become denser, while the number of targets remains limited. This leads to ambiguous situations, where multiple queries point to the same target. 
Such ambiguity often requires more time for convergence. Experimental results show that when we extended the schedule for the $N_q=30$ experiment to 2$\times$, performance improved further, indicating that a larger number of queries requires more iterations to converge. 
After balancing these considerations, we select $N_q=10$ as the default setting.

\subsubsection{Module Effectiveness Across Different Backbones}

To further validate the effectiveness and generalizability of our proposed approach, we conduct ablation experiments with three different backbone configurations: a dual-stream encoder based on CLIP pretraining (\textit{ViT-B}), a single-stream encoder (\textit{VILT-B}~\cite{vilt}), and the fusion encoder adopted in this work (\textit{BEiT3-B}~\cite{beit3}). For the ViT-B model, we additionally employ a 3-layer Transformer to fuse image and text features and obtain multi-modal representations. 
In all settings, the \textit{Modules} column refers to the inclusion of our proposed APD and PIPH modules. The default configuration follows a multi-task paradigm, where the detection branch adopts a DETR-style decoder and the segmentation branch follows the semantic segmentation formulation illustrated in Fig.~\ref{fig:framework}(c). 
As shown in Table~\ref{tab:ablation_backbone}, incorporating our modules yields consistent and notable performance gains across all backbone configurations under the generalized visual grounding setting. This consistency highlights that our approach is both backbone-agnostic and robust, demonstrating effectiveness across diverse architectural designs.

\begin{table}
  \makeatletter\def\@captype{table}
  \centering
  \footnotesize
  \renewcommand\arraystretch{1.2}
  \setlength{\tabcolsep}{2.0pt}
  \caption{Effectiveness of different backbones.}
  \vspace{-10pt}
  \resizebox{0.85\linewidth}{!}{
  \begin{tabular}{c|c|cccc}
  \specialrule{.1em}{.05em}{.05em} 
  Backbone        & Modules & F1score & N-acc. & gIoU & cIoU \\
  \hline
  CLIP-VIT-B~\cite{clip,vit}  &             & 59.33  & 64.01 & 60.24 & 56.20 \\
  CLIP-VIT-B~\cite{clip,vit} & \checkmark  & 63.12  & 68.99 & 65.01 & 60.98 \\
  \hline
  VILT-B~\cite{vilt}          &             & 64.88  & 68.75 & 65.93 & 63.81 \\
  VILT-B~\cite{vilt}          & \checkmark  & 68.72  & 72.32 & 70.01 & 65.79 \\
  \hline
  BEiT3-B~\cite{beit3}         &             & 67.13 & 69.98 & 67.33 & 64.90 \\
  BEiT3-B~\cite{beit3}         & \checkmark  & 71.43 & 75.87 & 72.41 & 67.39 \\
  \specialrule{.1em}{.05em}{.05em}
  \end{tabular}}
  \label{tab:ablation_backbone}
\end{table}

  \begin{figure*}
      \centering
      \includegraphics[width=0.95\linewidth]{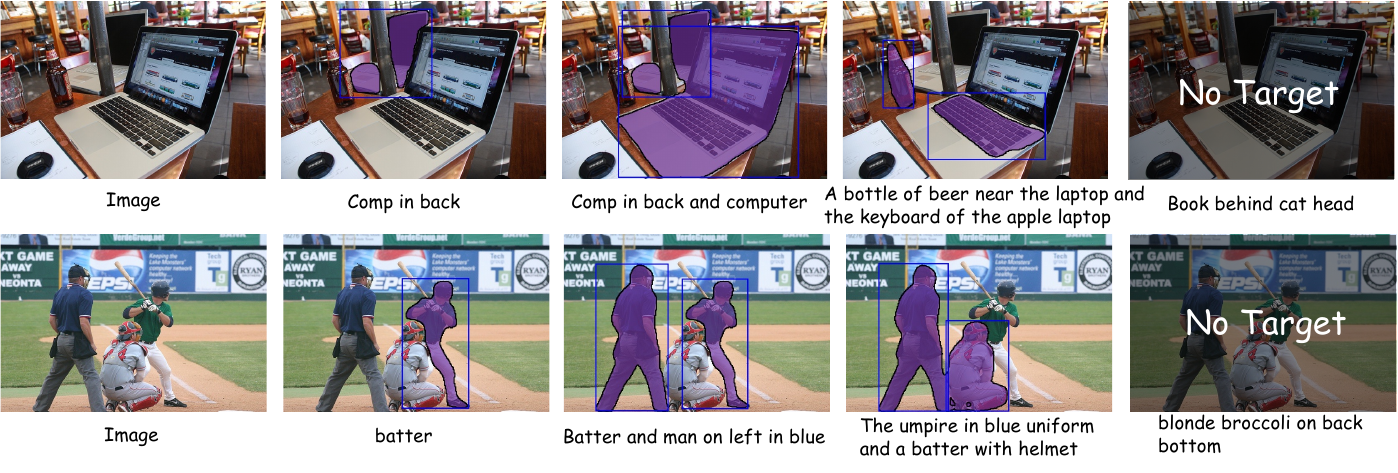}
      \vspace{-10pt}
      \caption{Visualization of multi-task predictions. Both GREC and GRES results for the same image under different expressions.}
      \label{fig:visual_pair}
  \end{figure*}
  
  \begin{figure*}
      \centering
      \includegraphics[width=1.0\linewidth]{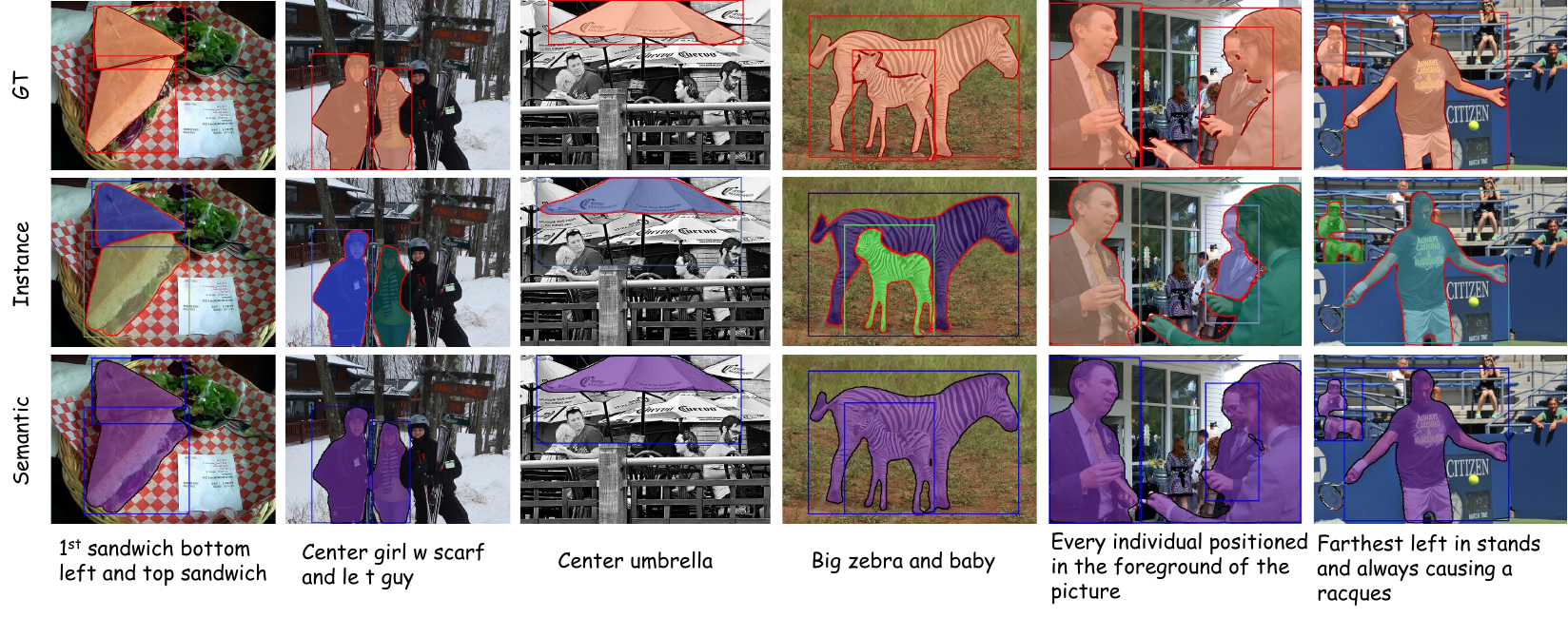}
      \vspace{-20pt}
      \caption{Instance-level Segmentation Results. The `Instance' row presents instance-level segmentation. The `Semantic' row presents the combination of both semantic segmentation and instance-level masks.}
      \vspace{-5pt}
      \label{fig:visual_instance}
  \end{figure*}

  \begin{figure*}
    \centering
    \includegraphics[width=1.0\linewidth]{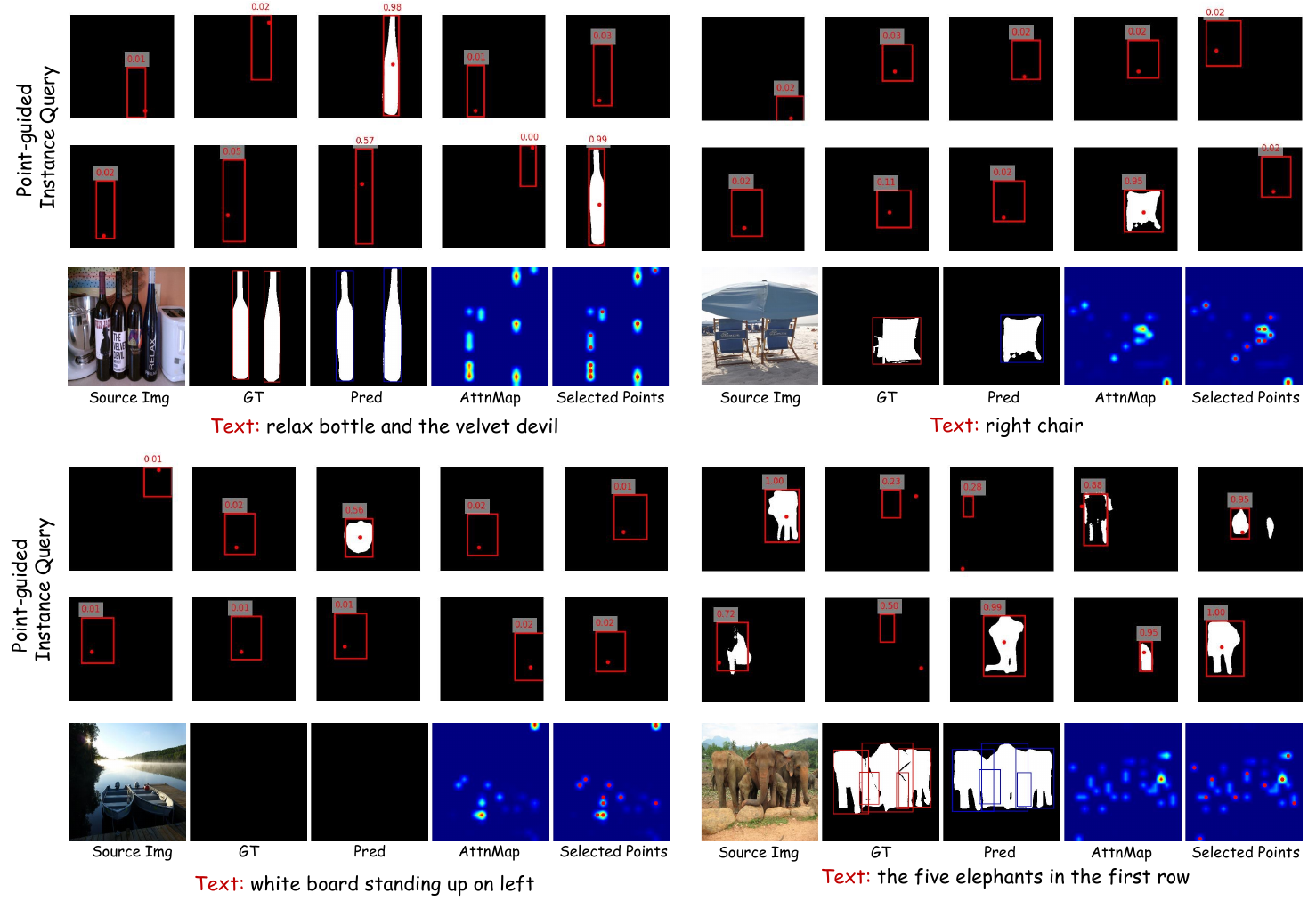}
    \vspace{-20pt}
    \caption{Visualization of intermediate process of point-guided instance-aware queries. The `Point-guided Instance Query' illustrates the points corresponding to each query, along with the predicted bounding boxes and masks. `AttnMap' represents the attention map from the IQG module, while `Selected Points' indicates the reference points output by the dynamic point selector.}
    \label{fig:visual_query}
  \end{figure*}


\subsubsection{Effectiveness of Post-Processing}
\label{subsubsec:ablation_postprocess}

\noindent{\bf The Impact of Score Threshold.}
Different loss function designs lead to varying prediction distribution trends. 
The outputs of InstanceVG tend to have higher confidence, as shown in Table~\ref{table:ablation_threshold}. When $thr_q$ increases from 0.7 to 0.9, both the F1score and gIoU improve. 
Based on this observation, we set the score threshold at $thr_q = 0.9$, which improves the F1score by 3.0\% and gIoU by 2.2\% compared to the score threshold of $thr_q = 0.7$.

\noindent{\bf The Impact of Mask Merge.} InstanceVG generates both global semantic- and instance-level segmentations. We analyzed the performance of these predictions both individually and when combined, as shown in Table~\ref{table:ablation_mask_output}. The instance-level predictions, which benefit from finer-grained supervision, achieve better performance compared to global predictions, improving gIoU by +1.6\%. Furthermore, merging the global and instance-level predictions yields an additional 0.5\% improvement in gIoU. The merging operation is performed using a logical OR, which helps preserve the integrity of the instance mask, resulting in better overall performance.

\noindent{\bf The Impact of NT Score and NMS.} As demonstrated in Table~\ref{table:ablation_nms}, we evaluated the effects of integrating the non-target (NT) branch's score into the query score and the influence of NMS. 
The introduction of the NT score effectively incorporates global confidence into each instance, resulting in a +1.2\% F1score and +0.7\% gIoU. 
However, the additional application of NMS has a minimal impact on the results, showing only slight improvement in detection while negatively affecting segmentation performance. 
This is because the design, which combines prior points with the original DETR matching paradigm, helps the model minimize the impact of ambiguous queries.

\subsubsection{Analysis of Parameters and Complexity}
\label{subsubsec:parameters}
We present a detailed incremental breakdown of the parameters and computational complexity of each module in InstanceVG in Table~\ref{table:params}. 
As can be observed, the combined parameter size of the proposed APD and PIPH modules is approximately 3M, accounting for only 1.6\% of the total model parameters. 
In terms of the computational complexity, APD and PIPH contribute 7.6\% of the overall computational cost, highlighting the efficiency of the proposed modules.

\section{Qualitative Results}
\label{sec:visualization}
In this section, we present the qualitative results of InstanceVG in four aspects: (1) visualization of multi-task generalized visual grounding (Sec.~\ref{subsec:visualization_mt}); (2) results for instance-level segmentation (Sec.~\ref{subsec:visualization_instance}); (3) visualization of the intermediate process of point-guided multi-task perception, including consistency predictions across points, boxes, and masks (Sec.~\ref{subsec:visualization_query}); (4) robustness visualization results in complex multi-referent target scenarios (Sec.~\ref{subsec:visualization_complex}).

\subsection{Visualization of Multi-task Predictions}
\label{subsec:visualization_mt}
InstanceVG effectively integrates and jointly accomplishes both the GREC and GRES tasks. 
As illustrated in Fig.~\ref{fig:visual_pair}, InstanceVG demonstrates synchronized execution of detection and segmentation, showcasing its capability to handle these tasks concurrently. 
In the visualization, various textual expressions are applied to the same image. 
As can be observed, InstanceVG exhibits strong robustness, even in scenarios involving the absence of targets or the presence of multiple referential targets.

\subsection{Visualization of Instance-aware Perception}
\label{subsec:visualization_instance}
Furthermore, InstanceVG demonstrates instance-aware capabilities, enabling fine-grained instance-level segmentation. 
As illustrated in Fig.~\ref{fig:visual_instance}, the instance-level segmentation results not only distinguish prominent objects (foreground semantics) but also accurately identify specific instances. This instance-aware capability allows InstanceVG to excel in multi-object scenarios. 
This performance is primarily attributed to the introduction of instance-level supervision, which equips the model with more comprehensive and enriched query representations. 
This supervision ensures consistent predictions across queries, boxes, and masks, thereby implicitly enhancing the model's alignment capability for diverse types of predictions.

\subsection{Visualization of Instance-aware Queries}
\label{subsec:visualization_query}
In Fig.~\ref{fig:visual_query}, we provide additional visualizations of InstanceVG's intermediate processes, including the points corresponding to the queries, the predicted boxes, and masks. 
It can be observed that InstanceVG achieves consistency across points, boxes, and masks for individual instances, with each point-guided query aligning consistently with the corresponding target. Furthermore, the attention maps from the IQG module and the corresponding selected points are also visualized. We can find that the prior points can effectively cover most potential instances.

\subsection{Visualization of Complex Multi-referent Target Scenarios}
\label{subsec:visualization_complex}
\begin{figure*}
  \centering
  \includegraphics[width=1.0\linewidth]{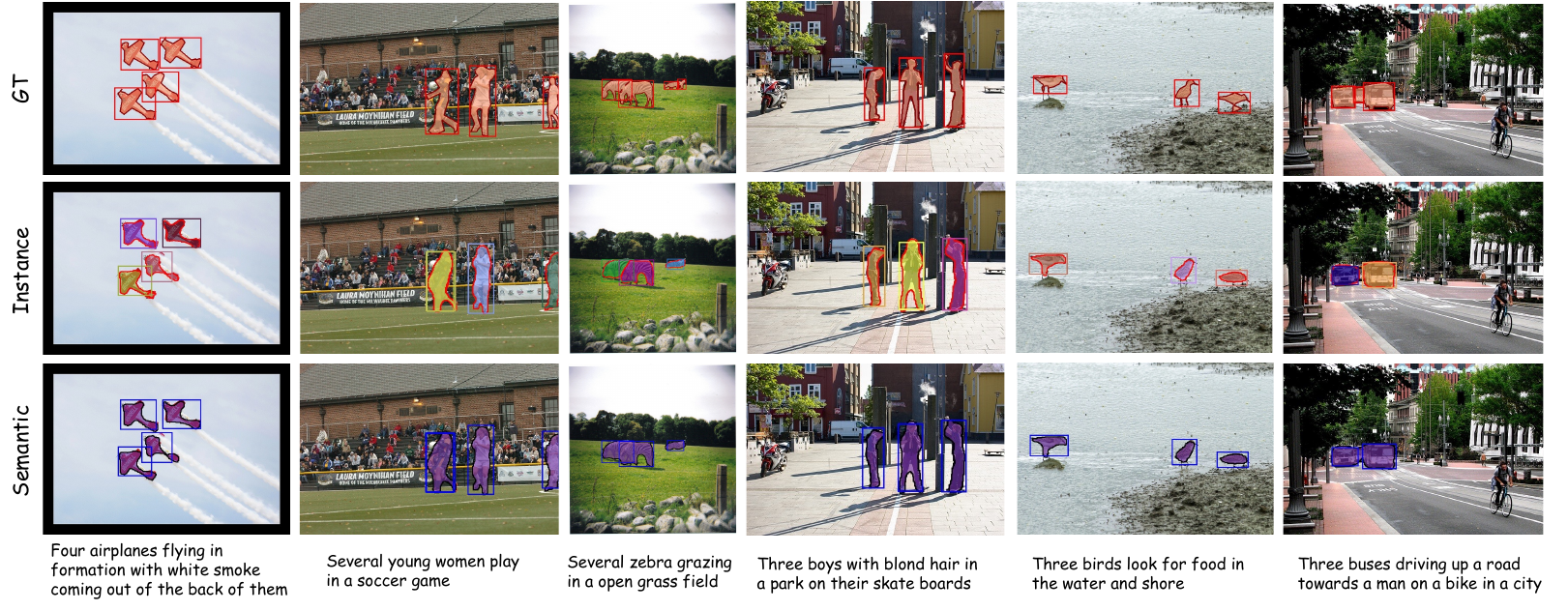}
  \vspace{-20pt}
  \caption{Visualization of complex and multi-referent objects situations in the Ref-ZOM~\cite{refzom} dataset. The `Semantic' row presents the combination of both semantic segmentation and instance-level masks.}
  \label{fig:visual_refzom}
\end{figure*}

In Fig.~\ref{fig:visual_refzom}, we present examples of complex multi-referent scenarios from the Ref-ZOM~\cite{refzom} dataset. While InstanceVG effectively perceives object locations, its detection accuracy for small objects remains limited, primarily due to constraints imposed by the model's input resolution.  
Moreover, the 16$\times$ downscaling of input resolution at the early stages reduces the model's ability to predict fine-grained contours. However, InstanceVG demonstrates a robust understanding of text-referred objects, maintaining high accuracy and recall even in complex multi-target scenarios.

\section{Conclusion and Future Work}
\label{sec:conclusion}

This paper introduces \textbf{InstanceVG}, an instance-aware multi-task generalized visual grounding framework that unifies the GREC and GRES tasks, while pioneering the exploration of instance-aware perception in generalized scenarios. 
To achieve these capabilities, we propose a novel attention-based point-prior decoder (APD) that adaptively selects prior reference points through attention maps, embedding spatial priors into queries to enhance instance-specific targeting. 
Additionally, we design a point-guided instance-aware perception head (PIPH), which facilitates the interaction between instance queries and global semantic features to generate instance-aware semantic queries, thereby establishing associations between queries, object boxes, and instance masks. 
The proposed InstanceVG framework demonstrates superior performance as compared with the existing methods, achieving state-of-the-art results across ten mainstream datasets spanning four distinct tasks (REC, RES, GREC, GRES).

However, there remain several areas for improvement: (1) The accuracy of target existence determination is suboptimal, highlighting the need for enhanced capabilities in cross-modal understanding of visual and textual inputs. (2) The model's ability to perceive small objects is limited, necessitating further exploration of how to leverage multi-level semantic features and coarse contour information effectively. (3) The fine-grained segmentation performance is inadequate due to the inherent limitations of ViT's patch embedding, which causes significant information loss during scale compression, resulting in coarse segmentation outputs. 

\section*{Acknowledgments}
This work is supported by the National Natural Science Foundation of China under Nos. 62276061 and 62436002.
This work is also supported by Research Fund for Advanced Ocean Institute of Southeast University (Major Program MP202404). This work is also supported by the SEU Innovation Capability Enhancement Plan for Doctoral Students (\text{CXJH\_SEU 25125}).

\bibliographystyle{IEEEtran}
\bibliography{InstanceVG.bib}

\end{document}